\pdfoutput=1

\documentclass[11pt]{article}

\usepackage[final]{acl}

\usepackage{times}
\usepackage{latexsym}

\usepackage[T1]{fontenc}

\usepackage[utf8]{inputenc}

\usepackage{microtype}

\usepackage{inconsolata}

\usepackage{graphicx}

\usepackage{enumerate}
\usepackage{amsmath}
\usepackage{amsfonts}
\usepackage{algorithm}
\usepackage{algorithmicx}
\usepackage{algpseudocode}
\usepackage{multirow}
\usepackage{booktabs}
\usepackage{makecell}
\usepackage{pifont}
\usepackage{tabularx}
\usepackage{array}
\usepackage{xcolor}
\usepackage{extarrows}
\usepackage[most]{tcolorbox}
\usepackage{subcaption}

\newcolumntype{L}{>{\raggedright\arraybackslash}X}
\newcolumntype{C}{>{\centering\arraybackslash}X}

\newcommand{\cmark}{\textcolor{green!70!black}{\ding{51}}} 
\newcommand{\xmark}{\textcolor{red}{\ding{55}}}            

\definecolor{lightgrassgreen}{RGB}{220, 240, 220}
\definecolor{lightblue}{RGB}{220, 230, 250}

\title{Beyond Spurious Signals: Debiasing Multimodal Large Language Models via Counterfactual Inference and Adaptive Expert Routing}

\author{
    Zichen Wu, 
    Hsiu-Yuan Huang,
    Yunfang Wu\thanks{~~Corresponding author.} \\
    School of Computer Science, Peking University \\
    MOE Key Laboratory of Computational Linguistics, Peking University\\
    National Key Laboratory for Multimedia Information Processing, Peking University \\ 
    \texttt{wuzichen@pku.edu.cn,  wuyf@pku.edu.cn}
}

\begin{document}
\maketitle
\begin{abstract}
Multimodal Large Language Models (MLLMs) have shown substantial capabilities in integrating visual and textual information, yet frequently rely on spurious correlations, undermining their robustness and generalization in complex multimodal reasoning tasks. This paper addresses the critical challenge of superficial correlation bias in MLLMs through a novel causal mediation-based debiasing framework. Specially, we distinguishing core semantics from spurious textual and visual contexts via counterfactual examples to activate training-stage debiasing and employ a Mixture-of-Experts (MoE) architecture with dynamic routing to selectively engages modality-specific debiasing experts. Empirical evaluation on multimodal sarcasm detection and sentiment analysis tasks demonstrates that our framework significantly surpasses unimodal debiasing strategies and existing state-of-the-art models. For further research, we release the training/evaluation pipelines at Github\footnote{\url{https://github.com/Zichen-Wu/Multimodal-Mixture-of-Expert-Debiasing.}}.
\end{abstract}

\section{Introduction}

Multimodal Large Language Models (MLLMs) have demonstrated significant capabilities in integrating information from various modalities, such as vision and language, achieving notable success in multimodal tasks~\citep{openai2024gpt4technicalreport, NEURIPS2023_6dcf277e, Chen_2024_CVPR, wang2024qwen2}. By unifying modalities, MLLMs can capture richer semantic information than unimodal approaches, establishing them as a promising direction for complex, real-world applications~\citep{wang2024comprehensivereviewmultimodallarge}.

\begin{figure}
    \centering
    \includegraphics[width=0.95\linewidth]{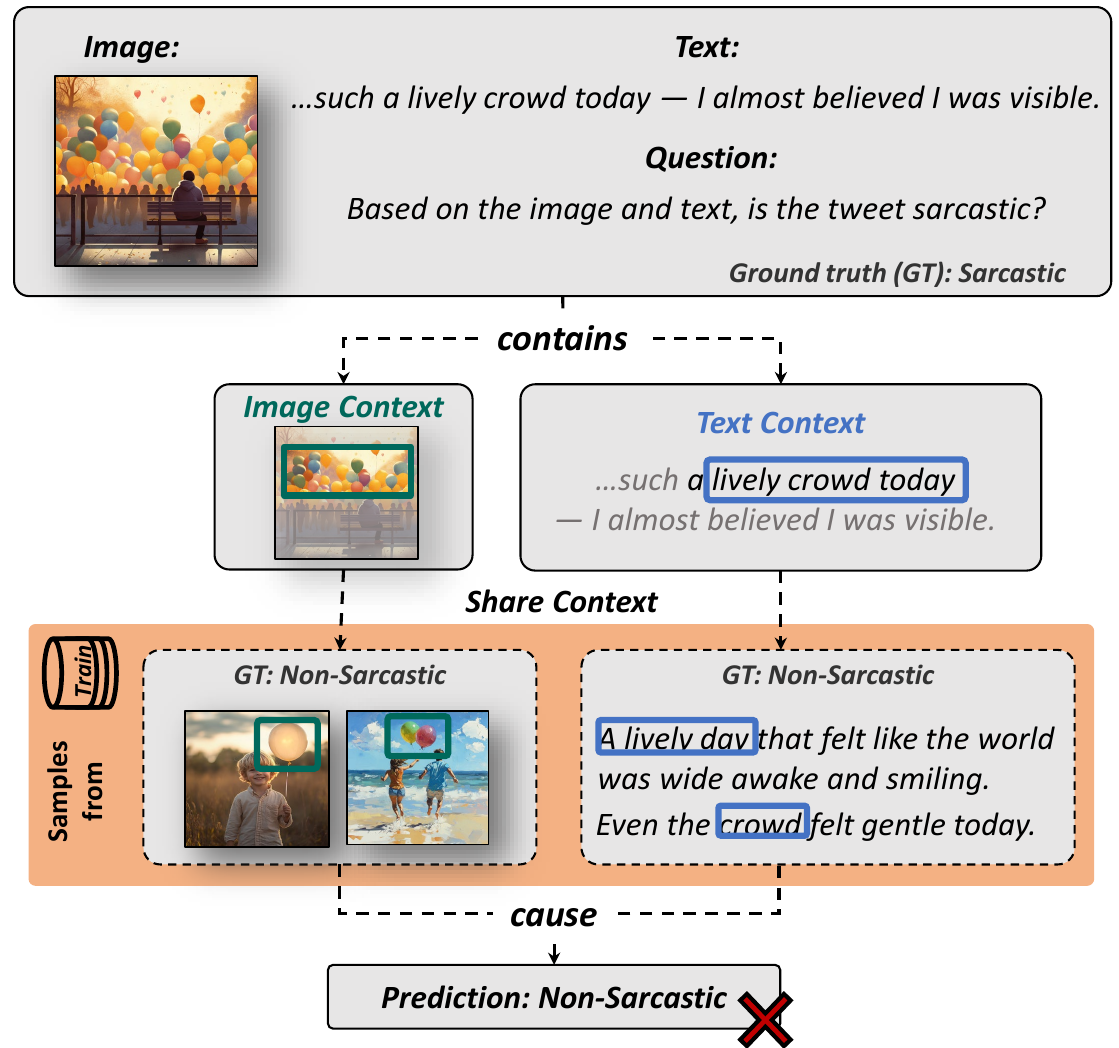}
    \caption{A diagram shows how a large model is negatively impacted by learning spurious correlations during training. In the depicted example, a non-sarcastic test sample (GT) is misclassified as sarcastic because its features frequently co-occurred with sarcastic labels in the training data.}
    \label{fig:multimodal-bias}
\end{figure}

\begin{figure}[t]
    \includegraphics[width=0.95\linewidth]{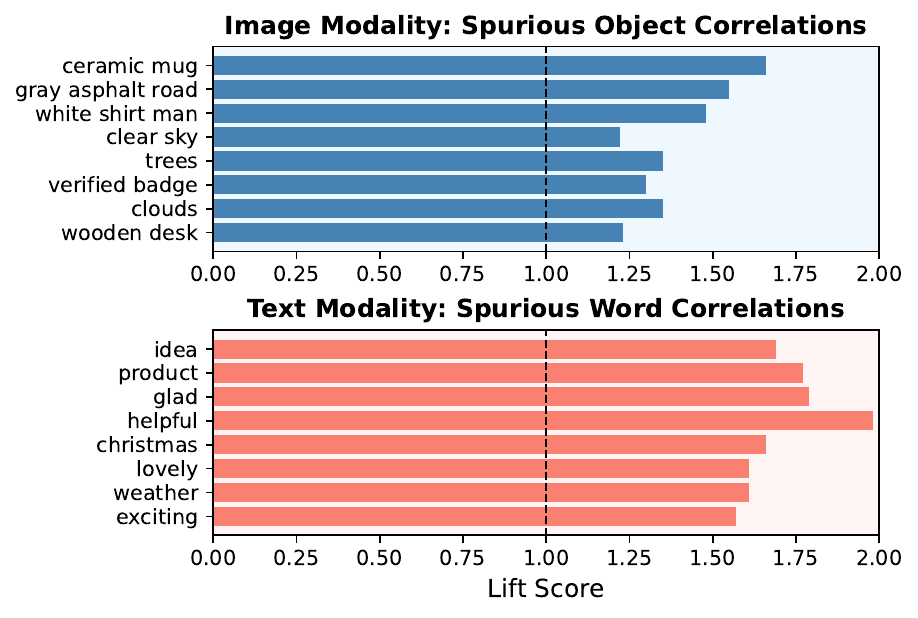}
    \caption{Lift scores for spurious correlations in image (objects) and text (words) modalities. Scores above 1.0 indicate positive spurious correlations.}
    \label{fig:spurious-correlation}
\end{figure}
Despite this progress, current MLLMs remain unreliable on tasks requiring nuanced semantic understanding and reasoning. In practice, they often over-rely on spurious correlations in one modality instead of truly fusing information, which leads to hallucinations and poor generalization~\citep{hosseini2025seeingwhatstherespurious, zhang2024debiasingmultimodallargelanguage}. Prior studies have found that these models may latch onto superficial cues present in the training data, leading to biases that are then amplified by large-scale pretraining~\citep{zhao2024lookingtextreducinglanguage}. For example, in sarcasm detection dataset MMSD2.0~\citep{qin-etal-2023-mmsd2}, certain words (e.g., ``weather'') or objects in an image (e.g., ``ceramic mug'') appeared frequently with the sarcastic label (Fig.~\ref{fig:spurious-correlation}), inadvertently cueing the model to predict ``sarcasm'' whenever it encounters those features. After supervised training, the model learns such spurious associations instead of genuine cross-modal reasoning, resulting in brittle performance on test data where the same shortcuts do not hold. These observations highlight the need for methods to make MLLMs focus on core semantics rather than incidental correlations.

To address this ``superficial correlation'' bias, causal mediation analysis~\citep{10.1145/3501714.3501734} offers a powerful framework for enhancing the reliability of multimodal semantic understanding. Its core principle involves modeling the multimodal task as a causal graphical model. By generating counterfactual inputs (isolating suspected biased elements) and comparing the model's predictions on these versus original inputs, it becomes possible to quantify and thereby mitigate spurious correlations between input features and the model's output.

However, existing approaches applying causality to MLLMs face limitations. Some methods perform debiasing operations on only a single modality~\citep{Niu_2021_CVPR, Agarwal_2020_CVPR, patil-etal-2023-debiasing} or apply corrections solely during the inference stage~\citep{ijcai2024p739, 10.1007/978-3-031-73636-0_27}, neglecting the significant impact of multi-modality and debiasing the model during the training. Other works focus on debiasing learned representations in a more general sense and have not been effectively adapted to modern MLLMs~\citep{pmlr-v235-zhang24as, chen2024multimodalsentimentanalysisbased}. These shortcomings often fail to comprehensively address biases arising from intricate multimodal interactions, leaving the model's internal representations suboptimal. Based on these observations, we pose the following research question: \textbf{How can causal mediation analysis be employed to jointly debias the superficial correlations within both textual and visual pathways of an MLLM, at both the training and inference stages?}

To address this issue, we first propose a multimodal causal analysis framework that explicitly distinguishes core semantic information from spurious contextual cues within each modality. Leveraging large pretrained models, we automatically extract superficial context from both textual and visual modalities to construct counterfactual samples. By penalizing model reliance on these counterfactual samples during training, we effectively integrate debiasing into the learning stage, thereby enhancing the robustness and representational power of the main multimodal model.

Recognizing that the automatically extracted superficial contexts may contain inaccuracies and that not all samples require uniform debiasing treatment, we further introduce a routing mechanism combined with a Mixture-of-Experts (MoE) architecture. Under this framework, dedicated expert models specifically handle the debiasing tasks for textual and visual modalities independently. The router dynamically learns and determines which debiasing expert(s) should be activated for the given sample, ensuring tailored and efficient debiasing. During training, these experts are exposed to modality-specific counterfactual data, compelling them to specialize in identifying and mitigating spurious influences. At inference time, the integrated system combines predictions from the main model and the appropriate bias experts, as guided by the router, to yield robust and accurate final outcomes.

We validate the proposed approach on two challenging multimodal semantic understanding tasks: sarcasm recognition (MSD) and sentiment analysis (MSA). Our experiments demonstrate that the causal debiasing framework substantially improves reliability and accuracy compared to both unimodal debiasing baselines and state-of-the-art task-specific models. Notably, the model achieves better generalization to different debiasing categories where naive finetuning falters, confirming that it learned to discount superficial correlations and focus on true multimodal semantics. These results underscore the benefits of a principled, joint debiasing strategy for MLLMs. 
In summary, our contributions include:
\begin{itemize}
\setlength\itemsep{0pt}
\setlength\parskip{0pt}
\setlength\topsep{0pt}
\item We formulate a multimodal debiasing approach grounded in causal mediation analysis, which simultaneously addresses both text and image biases.
\item We propose a novel expert-based architecture with a gating mechanism to isolate and remove spurious influences during both training and inference.
\item Our approach achieves state-of-the-art results on sarcasm and sentiment tasks, enhancing robustness and interpretability.
\end{itemize}
\section{Related Work}
\subsection{Bias in MLLMs}
Multimodal vision-language models often learn unintended spurious correlations between modality-specific cues and target outputs, causing biased predictions and poor generalization. For instance, Visual QA models frequently exploit dataset biases, such as responding \textit{yes} to questions starting with \textit{Do you see} due to learned shortcuts~\citep{Niu_2021_CVPR, Kim_2023_CVPR}. These biases may arise from coincidental visual or textual patterns. MLLMs trained on web-scale data can further inherit and amplify such modality-specific biases, which drives addressing these biases crucial for robust multimodal understanding~\citep{ye2024mmspubenchbetterunderstandingspurious, zhang2024debiasingmultimodallargelanguage}.

\subsection{Causal Mediation Analysis and Interventions}
Structural Causal Models (SCMs) offer a principled approach to explicitly model and mitigate spurious correlations by using interventions and counterfactual reasoning~\citep{10.1145/3501714.3501734}, a direction that has attracted increasing attention~\cite{foil2018, thrush2022winoground, le2023coco, zheng2024snse,leng2024vcd,zhu2024vspa}.

Recent studies apply causal inference frameworks in multimodal tasks to quantify and remove modality-induced biases \cite{palit2023vlm,golovanevsky2024notice,yu2024rlhfv}. For instance, \citet{ijcai2024p739} used counterfactual text generation to address textual biases in sarcasm detection, and \citet{10.1007/978-3-031-73636-0_27} adopted a similar approach for sentiment analysis. \citet{chen2024multimodalsentimentanalysisbased} extended these strategies by explicitly modeling biases in both image and text modalities, but did not investigate the bias originating from spurious parts in a fine-grained manner. In VQA tasks, causal reasoning methods include the Chain-of-Thought (CoT)-based CAVE module by \citet{chen-etal-2024-quantifying}, and \citet{liu-etal-2024-ce}'s entropy-based counterfactual debiasing strategy for video-grounded QA. Additionally, \citet{Yang_2024_CVPR} applied causal inference in emotion recognition by decomposing context into relevant and irrelevant cues, eliminating distracting signals during inference.
Unlike previous methods, our proposed approach explicitly constructs fine-grained counterfactual samples, integrating causal debiasing directly into both training and inference to comprehensively address multimodal biases.

\begin{figure}[t]
    \centering
    \includegraphics[width=\linewidth]{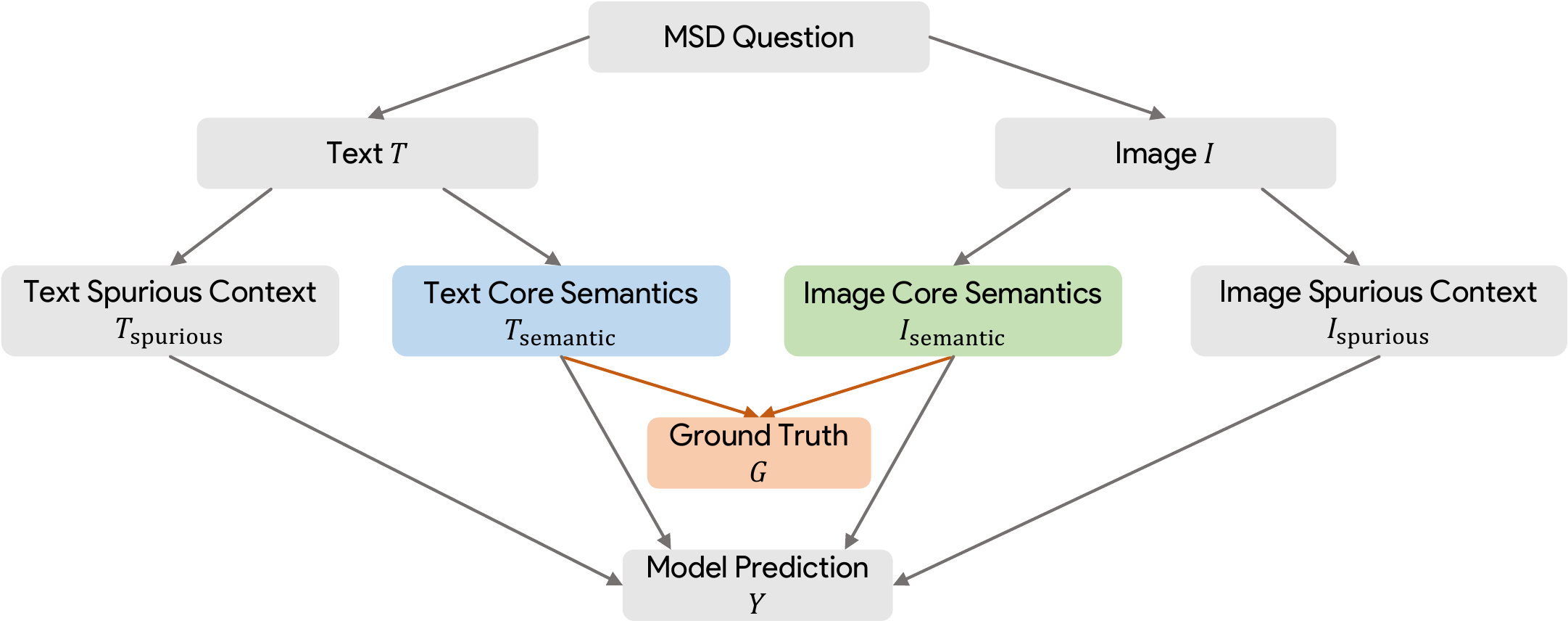}
    \caption{Causal Graph for Multimodal Sarcasm Detection. The non-grey regions indicate the ideal causal mechanisms. Unbiased prediction results could be achieved by separating image and text inputs into semantic and spurious components and performing controlled interventions.}
    \label{fig:multimodal-causal-graph}
\end{figure}
\section{Preliminaries}
\subsection{Multimodal Causal Mediation Framework}
\label{sec:multimodal-causal-graph}
MLLMs often suffer biases from spurious textual and visual contexts, degrading prediction accuracy. We propose a multimodal causal mediation framework to explicitly model causal relationships among multimodal inputs, bias-mediating variables, and outputs.

Taking sarcasm detection as an example, inputs include text ($T$) and image ($I$), ideally generating semantic features ($T_{semantic}$, $I_{semantic}$). However, irrelevant contextual features ($T_{spurious}$, $I_{spurious}$) may bias outcomes. We illustrate these effects via a multimodal causal graph (Fig.~\ref{fig:multimodal-causal-graph}), distinguishing:
\par\noindent\textbf{Unbiased path:} $T \to T_{semantic} \to Y$ and $I \to I_{semantic} \to Y$, capturing desired accurate semantic understanding.
\par\noindent\textbf{Biased path:} $T \to T_{spurious} \to Y$ and $I \to I_{spurious} \to Y$, capturing biases introduced by spurious information.

We capture the Natural Direct Effect (NDE) of inputs on the prediction outcome as the unbiased results. In causal mediation theory, it refers to isolating the direct influence of the relevant semantic information ($T_{semantic}, I_{semantic}$), while holding mediating biases constant at baseline levels. Based on the multimodal causal graph, we quantify these effects through three scenarios:
\par\noindent\textbullet{ Textual Counterfactual Scenario:} Isolate $T_{spurious}$ and mask visual input:
\begin{equation}
  Y_t = Y(T_{spurious}, \phi).
\end{equation}
\par\noindent\textbullet{ Visual Counterfactual Scenario:}  Isolate $I_{spurious}$ and mask textual input:
\begin{equation}
    Y_i = Y(\phi, I_{spurious}).
\end{equation}
\par\noindent\textbullet{ Original Scenario:} Use the full original input:
\begin{equation}
   Y_0 = Y(T, I).
\end{equation}

We then extract the unbiased output by computing the difference between the original and counterfactual predictions:
\begin{equation}
    Y_{\text{unbiased}} = \mathrm{DIFF}(Y_0, Y_t, Y_i),
\end{equation}
This provides a principled estimate of prediction free from modality-specific spurious bias.


\subsection{Counterfactual Content Construction}
Building upon our multimodal causal mediation framework (Sec.~\ref{sec:multimodal-causal-graph}), the ability to isolate specific causal factors necessitates the construction of targeted counterfactual contents. Unlike traditional methods relying on heuristic rules or dataset annotations, we automate the generation of fine-grained multimodal counterfactual inputs ($T_{spurious}, I_{spurious}$) using MLLMs. For textual inputs, this involves identifying and masking core semantic segments. For visual inputs, we utilize attention mechanisms to pinpoint and modify salient regions. Due to space constraints, a detailed methodological description is provided in App.~\ref{app:counterfactual_content_construction}. These meticulously constructed counterfactuals are instrumental for the causal debiasing techniques presented in following sections.

\section{Multimodal Debiasing Methods}
In this section, we propose methods to mitigate spurious multimodal biases, targeting both inference and training stages:
\par\noindent1. \textbf{Multimodal Inference Debiasing (MID):} An inference-stage, plug-and-play method for external bias correction without altering model parameters.
\par\noindent2. \textbf{Multimodal Training Debiasing:} To embed robust representations, we integrate debiasing directly into model training. This involves foundational principles of counterfactual-aware training. which uses counterfactual samples to reduce spurious correlations. Building upon these, we introduce our primary training-stage method, \textbf{Multimodal Mixture-of-Experts Joint Debiasing (MME-JD)}. This advanced MoEs approach uses a learned router to adaptively dispatch samples to specialized expert branches, enabling fine-grained debiasing.
These methods are detailed in the subsequent sections.

\subsection{MID: Multimodal Inference Debiasing}
\label{sec:mid}
The simplest debiasing approach applies only at inference, without any additional training procedure. Leveraging the proposed multimodal causal mediation anlysis framework, we could easily adapt it to MLLMs by using counterfactual samples in an inference-time plug-and-play manner.

Given original multimodal inputs (image $I$, text $T$), we first derive the original prediction probabilities $p_0$. We then generate counterfactual samples icluding text-only context ($\hat{T}$) and image-only context ($\hat{I}$), to isolate modality-specific biases, yielding probability distributions $p_t$ and $p_i$, respectively. Each $\{ p_0, p_{t}, p_{i} \}$ is a vector of length $K$ (for a $K$-class prediction task).

To suppress the bias introduced by spurious text and/or image context, we perform a linear correction on $p_0$:

\vspace{-0.7em}
\small
\begin{equation}
\label{eq:mid}
    \tilde{p} = p_0 - \alpha p_i - \beta p_t,
\end{equation}
\normalsize
where $\alpha$ and $\beta$ are hyperparameters controlling how aggressively we subtract the counterfactual probabilities from the original prediction.

Choosing $\alpha, \beta$ is non-trivial: different datasets and tasks may require stronger or weaker correction. We follow a validation-set-based approach, searching over $\alpha,\beta \in [0,1]$ to maximize a performance metric Eval (We use F1 in experiments):

\small
\begin{equation}
    \hat{\alpha}, \hat{\beta} = \arg\max_{\alpha,\beta\in[A,B]}  Eval(D|f, \alpha, \beta).
\end{equation}
\small
\normalsize
We employ Bayesian optimization to efficiently find the best-performing parameters.
\begin{figure*}[!ht]
    \centering
    \includegraphics[width=0.9\linewidth]{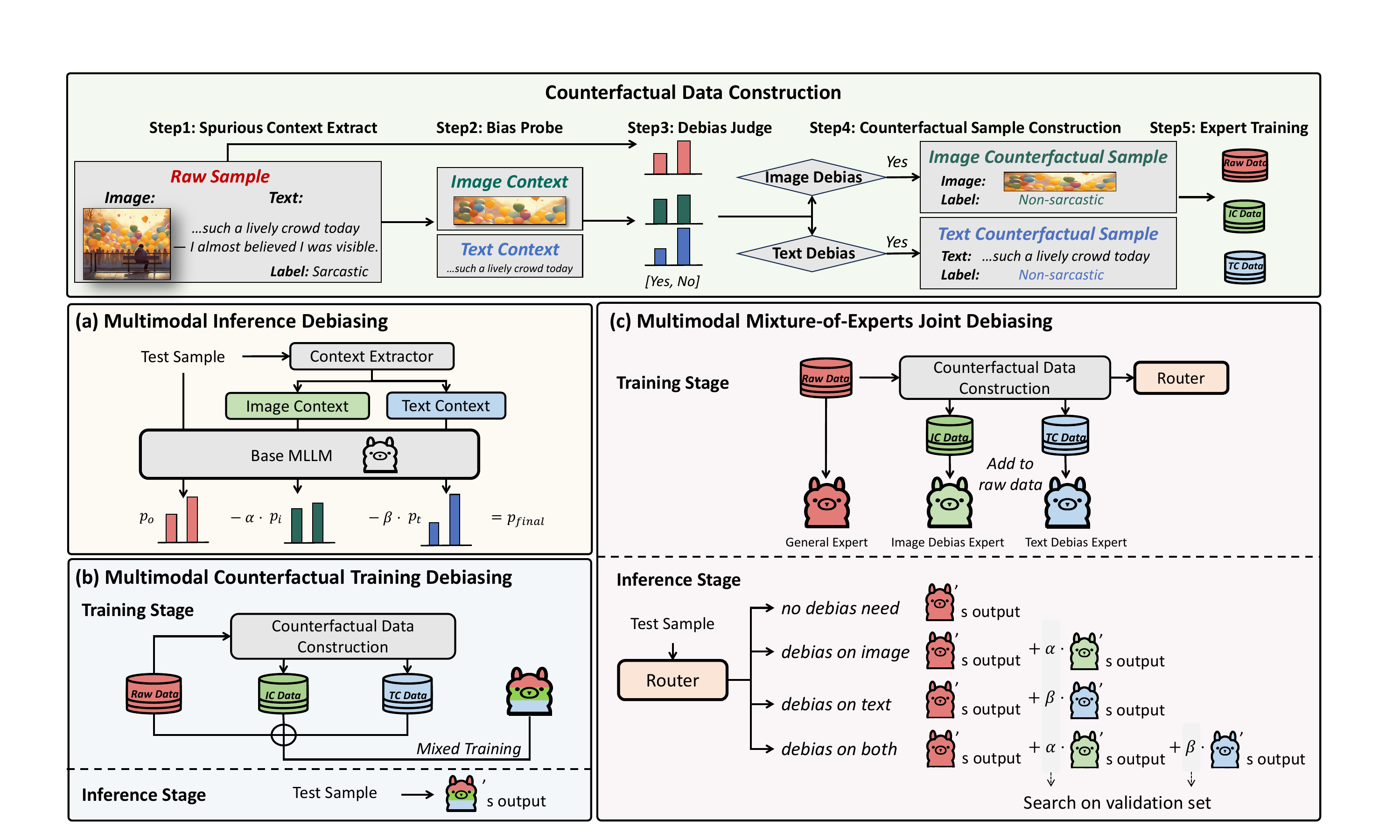}
    \caption{Overview of proposed multimodal debiasing frameworks: (a) inference-only debiasing using context extraction, (b) counterfactual training debiasing via mixed data augmentation, and (c) Mixture-of-Experts joint debiasing incorporating a dynamic router mechanism.}
    \label{fig:overall-method}
\end{figure*}

\subsection{Intergate Causal Debiasing into Training}
While MID provides a practical inference-time fix, it operates externally and does not fundamentally alter the learned representations or the model's reliance on spurious features. To encourage the model to learn representations that are intrinsically more robust to spurious multimodal correlations, we shift our focus from inference-time correction to integrating counterfactual information directly into the training process.
\subsubsection{Foundations}
\label{sec:foundations}
A direct approach to incorporate the causal effect of spurious contexts is to model a bias-removed prediction (Eq.~\ref{eq:mid}) into the loss function:

\vspace*{-1em}\small
\begin{equation}
    \mathcal{L} = -\mathbb{E}[\log \text{norm}(P(y|i, t) - \alpha \cdot  P(y|\hat{i}) - \beta \cdot P(y|\hat{t}))],
\end{equation}
\normalsize
where $\text{norm}(\cdot)$ is a softmax normalization. However, this approach is computationally expensive (three forward passes per step) and can suffer from numerical instability due to its internal subtraction.

To address these issues while maintaining the same fundamental training goals, we propose an alternative strategy centered on constructing counterfactual training objectives using reversed labels. The core objective remains consistent with $\mathcal{L}$ to maximize $P(y|i,t)$ on original inputs while minimizing reliance on spurious contexts. This is practically implemented by encouraging the model to predict an incorrect label $\hat{y}$, when presented with only the spurious context. This leads to the following training objective, $\mathcal{L}'$:

\vspace*{-1em}\small
\begin{align}
    \mathcal{L}' &= \underbrace{-\mathbb{E}[\log P(y|i, t)]}_{\text{(I) Maximize original accuracy}} + \underbrace{\mathbb{E}[\log P(y|\hat{i}) + \log P(y|\hat{t})]}_{\text{(II) Penalize spurious reliance via $y$}} \label{eq:l_prime_formulation1} \\
    &\xlongequal{\text{approx.}} \text{(I)} \underbrace{- \mathbb{E}[\log P(\hat{y}|\hat{i})] - \mathbb{E}[\log P(\hat{y}|\hat{t})]}_{\text{(III) Minimize spurious accuracy via $\hat{y}$ (reversed label)}}. \label{eq:l_prime_reversed_label}
\end{align}
\normalsize
This formulation provides an efficient way for training-time debiasing by incorporating counterfactual samples into the training data, requiring only a single forward pass per instance and thus enhancing efficiency.

While $\mathcal{L}'$ (Eq.~\ref{eq:l_prime_reversed_label}) enables counterfactual training, MLLMs may hallucinate or misattribute spurious correlations. Indiscriminate use of naively generated counterfactuals can therefore introduce noise. To selectively identify informative samples for debiasing, we compare the original prediction $p_0(y)$ with counterfactual predictions $p_t(y), p_i(y)$  obtained from the base model (as in Sec.~\ref{sec:mid}) on truth $y$ using a tolerance $\varepsilon$:

\noindent a. \textbf{No debiasing}: $p_t(y) > p_0(y) + \varepsilon \land p_i(y) > p_0(y) + \varepsilon$ (i.e., both contexts aid prediction).
\par\noindent b. \textbf{Visual debiasing only}: $p_i(y) + \varepsilon < p_0(y) < p_t(y) - \varepsilon$ (i.e., visual context harms, text aids/is neutral). Construct visual counterfactual.
\par\noindent c. \textbf{Textual debiasing only}: $p_t(y) + \varepsilon < p_0(y) < p_i(y) - \varepsilon$ (i.e., textual context harms, visual aids/is neutral). Construct textual counterfactual.
\par\noindent d. \textbf{Debias both modalities}: $p_t(y) + \varepsilon < p_0(y) \land p_i(y) + \varepsilon < p_0(y)$ (i.e., both contexts harm). Construct both counterfactuals.
\par\noindent e. \textbf{Exclude}: All other cases (e.g., within uncertainty margin).

A smaller $\varepsilon$ includes more samples for debiasing but can increase ambiguity near decision boundaries. We use $\varepsilon=0.1$ in our experiments.

\subsubsection{Training-Stage Only Debiasing}
\label{sec:mctd}
Building upon $\mathcal{L}'$ and counterfactual sample selection strategy, we could establish a specific training-time debiasing approach, \textbf{MCTD} (\textbf{Multimodal Counterfactual Training Debiasing}), illustrated in Fig.~\ref{fig:overall-method}~(b). During inference, MCTD directly accepts test samples $(i,t)$ to produce outputs, requiring no post-hoc adjustment.\footnote{We include MCTD as a distinct method here for experimental comparison, allowing us to assess its training-stage debiasing performance against inference-stage approaches.}

\subsection{MME-JD: Multimodal Mixture-of-Experts Joint Debiasing}
While the basic Counterfactual Training (Sec.\ref{sec:mctd}) improves internal representation robustness, its uniform application of counterfactual augmentation may not be ideal for all samples. To accommodate diverse bias types and intensities with sample-specific treatment, we propose MME-JD, a Multimodal Mixture-of-Experts Joint Debiasing approach, where each sample dynamically selects appropriate debiasing experts through a learned routing mechanism (as in Fig.~\ref{fig:overall-method}~(c)).

\subsubsection{Architecture and Objectives}
We employ three parallel expert branches:
\par\noindent 1. General Expert, trained conventionally on original samples:

\vspace{-1em}
\small
\begin{equation}
    \mathcal{L}_{\mathrm{GE}} = - \mathbb{E}[\log P(y|i, t)].
\end{equation}
\normalsize
\par\noindent 2. Image Debiasing Expert, expliciting incorporating counterfactual visual examples selected by Sec~\ref{sec:mctd} criterion b,d into training:

\small
\begin{equation}
    \mathcal{L}_{\mathrm{IDE}} = - \mathbb{E}[\log P(y|i, t)] - \mathbb{E}[\log P(\hat{y}|\hat{i})].
\end{equation}
\normalsize
\par\noindent 3. Text Debiasing Expert (TDE), analogously reducing textual bias via:

\small
\begin{equation}
    \mathcal{L}_{\mathrm{TDE}} = - \mathbb{E}[\log P(y|i, t)] - \mathbb{E}[\log P(\hat{y}|\hat{t})].
\end{equation}
\normalsize

\subsubsection{Router and Inference-time Combination}
\label{sec:counterfactual-sample-categorize}
A key component in MME-JD is a router module designed to dynamically determine the most suitable expert combination for any given input sample. The router takes the original and counterfactual inputs $(i,t,\hat{i},\hat{t})$ and is trained as a classifier to predict an optimal expert strategy label, $c$, for each training instance. This strategy label $c$ is assigned by applying the heuristic counterfactual sample selection criteria detailed previously in Sec.~\ref{sec:foundations}. We assign $c \in \{0,1,2,3\}$ corresponding to the required expert combination: \textit{GE only} (0), \textit{GE + IDE} (1), \textit{GE + TDE} (2), and \textit{GE + IDE + TDE} (3).

\label{sec:router}
The Router is trained as a classifier to predict the expert strategy label c for each sample. Its training objective is to minimize the cross-entropy loss:

\small
\begin{equation}
    \mathcal{L}_{routing} = -\mathbb{E} \log Q(c|i,t,\hat{i},\hat{t}).
\end{equation}
\normalsize

During inference, for a given sample $(i,t)$, after generating $\hat{t}, \hat{i}$, the router first predicts the optimal expert strategy by

\small
\begin{equation}
    c^* = \arg\max Q(c|i,t,\hat{i},\hat{t}).
\end{equation}
\normalsize
Then, the relevant experts (GE, and TDE/IDE if selected by $c^*$) process their respective inputs to produce outputs $p_{\text{GE}}, p_{\text{TDE}}, p_{\text{IDE}}$. The final output logits are computed by combining the outputs of the selected experts based on the strategy $c^*$.

\vspace{-0.5em}
\small
\begin{align}
\tilde{p} \;=\;
\begin{cases}
p_{\mathrm{GE}}, &  c^*=0 \\
p_{\mathrm{GE}} + \alpha_1\, p_{\mathrm{IDE}}, &  c^*=1 \\
p_{\mathrm{GE}} + \beta_2\, p_{\mathrm{TDE}}, &  c^*=2 \\
p_{\mathrm{GE}} + \alpha_3\, p_{\mathrm{IDE}} + \beta_3\, p_{\mathrm{TDE}}. &  c^*=3
\end{cases}
\end{align}
\normalsize
where $\alpha_{c}$ and $\beta_{c}$ are scalar weighting parameters that could be searched on validation set as Sec.~\ref{sec:mid}.

\subsection{Computational Overhead}

Table~\ref{tab:overhead} reports relative training and inference costs (normalized to the base model). Inference cost mainly reflects the number of forward passes over the VLM: TFCD adds one textual counterfactual ($\sim$2$\times$), MID adds two counterfactuals (3$\times$). MCTD folds counterfactual supervision into training, keeping inference at 1$\times$; MME-JD trains experts and routes at test time, yielding $\sim$3$\times$ for both training and inference.

\begin{table}[t]
\centering
\small
\begin{tabular}{lcc}
\toprule
\textbf{Method} & \textbf{Training Cost} & \textbf{Inference Cost} \\
\midrule
\textbf{Base Model} & 1$\times$ & 1$\times$ \\
\textbf{TFCD} (text debias only) & 1$\times$ & $\sim$2$\times$ \\
\textbf{MID} & 1$\times$ & 3$\times$ \\
\textbf{MCTD} & $\sim$1$\times$ & 1$\times$ \\
\textbf{MME-JD} & $\sim$3$\times$ & 3$\times$ \\
\bottomrule
\end{tabular}
\caption{Relative computational overhead vs.\ the base model.}
\label{tab:overhead}
\end{table}

\begin{table*}[!ht]
\centering
\small
\begin{minipage}{0.9\textwidth}
\centering
\small
\begin{tabularx}{\textwidth}{@{\extracolsep{\fill}}l*{4}{c}@{}}
\toprule
\textbf{Method} & \textbf{Acc.} & \textbf{Prec.} & \textbf{Recall} & \textbf{F1} \\
\midrule
HFM & 71.04 $\pm$ 0.29 & 64.92 $\pm$ 1.36 & 69.63 $\pm$ 0.93 & 67.01 $\pm$ 0.86 \\
Att-BERT & 80.10 $\pm$ 1.34 & 76.35 $\pm$ 2.73 & 77.76 $\pm$ 0.54 & 77.14 $\pm$ 1.46 \\
CMGCN & 79.92 $\pm$ 1.40 & 75.84 $\pm$ 1.16 & 78.10 $\pm$ 1.78 & 76.86 $\pm$ 1.45 \\
HKE & 76.47 $\pm$ 1.31 & 73.51 $\pm$ 1.00 & 71.62 $\pm$ 2.62 & 72.40 $\pm$ 1.72 \\
Multi-view CLIP & 85.35 $\pm$ 0.37 & 81.37 $\pm$ 1.12 & 87.05 $\pm$ 0.60 & 83.28 $\pm$ 1.10 \\
\midrule
InternVL2.5 & 85.76 $\pm$ 0.57 & 79.91 $\pm$ 1.51 & 89.39 $\pm$ 0.28 & 84.38 $\pm$ 0.61 \\
InternVL2.5 + TFCD & 85.96 $\pm$ 0.34 & 80.00 $\pm$ 0.69 & 89.87 $\pm$ 1.02 & 84.65 $\pm$ 0.45 \\
InternVL2.5 + MID & 86.34 $\pm$ 0.59 & 80.62 $\pm$ 1.75 & 89.88 $\pm$ 0.63 & 85.00 $\pm$ 0.45 \\
InternVL2.5 + MCTD & 86.26 $\pm$ 0.39 & 80.33 $\pm$ 1.03 & 90.16 $\pm$ 0.51 & 84.96 $\pm$ 0.25 \\
InternVL2.5 + MME-JD & \textbf{86.84} $\pm$ 0.41 & \textbf{81.25} $\pm$ 1.06 & \textbf{90.26} $\pm$ 1.04 & \textbf{85.52} $\pm$ 0.41 \\
\midrule
Qwen2-VL & 86.74 $\pm$ 0.29 & 81.14 $\pm$ 0.88 & 90.45 $\pm$ 0.81 & 85.54 $\pm$ 0.32 \\
Qwen2-VL + TFCD & 87.55 $\pm$ 0.35 & 82.70 $\pm$ 0.72 & 89.87 $\pm$ 0.53 & 86.14 $\pm$ 0.26 \\
Qwen2-VL + MID & 87.68 $\pm$ 0.51 & 83.14 $\pm$ 1.01 & 89.52 $\pm$ 0.84 & 86.21 $\pm$ 0.47 \\
Qwen2-VL + MCTD & 88.13 $\pm$ 0.34 & 82.51 $\pm$ 1.15 & \textbf{91.90} $\pm$ 0.62 & 86.95 $\pm$ 0.41 \\
Qwen2-VL + MME-JD & \textbf{88.42} $\pm$ 0.16 & \textbf{83.19} $\pm$ 0.80 & 91.61 $\pm$ 0.90 & \textbf{87.20} $\pm$ 0.23 \\
\bottomrule
\end{tabularx}
\caption{Comparison of our proposed methods (including MID, MCTD and MME-JD) with existing methods on dataset MMSD2.0.}
\label{tab:mmsd}
\end{minipage}
\begin{minipage}{0.9\textwidth}
\centering
\small
\begin{tabularx}{\textwidth}{@{\extracolsep{\fill}}l*{3}{c}@{}}
\toprule
\textbf{Method} & \textbf{Acc.} & \textbf{M-F1} & \textbf{W-F1} \\
\midrule
MVAN & 66.06 $\pm$ 0.97 & 54.45 $\pm$ 1.02 & 64.01 $\pm$ 1.12 \\
MGNNS & 67.49 $\pm$ 0.31 & 54.74 $\pm$ 1.72 & 64.37 $\pm$ 0.80 \\
CLMLF & 66.80 $\pm$ 0.71 & 54.93 $\pm$ 1.39 & 64.63 $\pm$ 0.89 \\
MDSE & 66.82 $\pm$ 1.26 & 55.12 $\pm$ 3.25 & 64.77 $\pm$ 0.92 \\
CF-MSA & 67.12 $\pm$ 1.28 & 55.18 $\pm$ 1.14 & 64.92 $\pm$ 1.01 \\
\midrule
InternVL2.5 & 71.02 $\pm$ 0.35 & 58.37 $\pm$ 1.36 & 69.66 $\pm$ 0.35 \\
InternVL2.5 + MCIS & 71.08 $\pm$ 0.54 & 59.23 $\pm$ 0.81 & 70.00 $\pm$ 0.53 \\
InternVL2.5 + MID & 71.52 $\pm$ 0.95 & 60.20 $\pm$ 1.24 & 70.60 $\pm$ 0.84 \\
InternVL2.5 + MCTD & 71.52 $\pm$ 0.75 & 60.08 $\pm$ 1.30 & 70.52 $\pm$ 1.04 \\
InternVL2.5 + MME-JD & \textbf{71.86} $\pm$ 0.76 & \textbf{60.64} $\pm$ 1.02 & \textbf{70.87} $\pm$ 0.68 \\
\midrule
Qwen2-VL & 69.79 $\pm$ 0.44 & 61.15 $\pm$ 0.71 & 69.98 $\pm$ 0.31 \\
Qwen2-VL + MCIS & 71.02 $\pm$ 0.47 & 60.35 $\pm$ 0.58 & 70.43 $\pm$ 0.33 \\
Qwen2-VL + MID & 71.46 $\pm$ 0.87 & 60.68 $\pm$ 0.93 & 70.92 $\pm$ 0.52 \\
Qwen2-VL + MCTD & 70.79 $\pm$ 0.66 & 60.55 $\pm$ 1.40 & 70.77 $\pm$ 0.73 \\
Qwen2-VL + MME-JD & \textbf{72.08} $\pm$ 0.46 & \textbf{62.42} $\pm$ 0.69 & \textbf{71.95} $\pm$ 0.35 \\
\bottomrule
\end{tabularx}
\caption{Comparison of our proposed methods with existing methods on MVSA-Multi. M-F1 denotes Macro-F1, while W-F1 refers to Weighted-F1.}
\label{tab:mvsa}
\end{minipage}
\end{table*}

\begin{table}[!ht]
  \centering
  \setlength{\tabcolsep}{0.9mm}
  \small
  \begin{tabularx}{\linewidth}{llllCCCC}
    \toprule
    \multicolumn{1}{l}{\multirow{2}{*}{\textbf{Methods}}} & \multicolumn{1}{l}{\multirow{2}{*}{\textbf{R.}}} & \multicolumn{1}{l}{\multirow{2}{*}{\textbf{C.}}} & \multicolumn{1}{l}{\multirow{2}{*}{\textbf{M.}}} & \multicolumn{2}{c}{\textbf{MMSD2.0}} & \multicolumn{2}{c}{\textbf{MVSA-Multi}} \\ \cmidrule(l){5-8} 
    \multicolumn{1}{c}{} &  &  &  & \textbf{Acc.} & \textbf{F1} & \textbf{Acc.} & \textbf{M-F1} \\ 
    \midrule
    \multicolumn{8}{c}{\textit{Inference-only}} \\ 
    \midrule
    Qwen2-VL        &     &     &     & 71.94 & 65.89 & 67.95 & 49.28 \\
    \; + MID     & \xmark & \xmark & \xmark & 73.22 & 66.97 & 69.12 & 51.41 \\
    \; + MRID    & \cmark & \xmark & \xmark & \textbf{74.88} & \textbf{68.21} & \textbf{69.29} & \textbf{52.83} \\
    \midrule
    \multicolumn{8}{c}{\textit{Training backbone}} \\ 
    \midrule
    Qwen2-VL        &     &     &     & 86.74 & 85.54 & 69.79 & 61.15 \\
    \; + MID     & \xmark & \xmark & \xmark & 87.68 & 86.21 & 71.46 & 60.68 \\
    \; + MRID    & \cmark & \xmark & \xmark & 88.09 & 86.58 & \textbf{72.58} & 61.94 \\
    \; + MCTD     & \xmark & \cmark & \xmark & 88.13 & 86.95 & 70.79 & 60.55 \\
    \; + MME-JD  & \cmark & \cmark & \cmark & \textbf{88.42} & \textbf{87.20} & 72.08 & \textbf{62.42} \\
    \bottomrule
  \end{tabularx}
  \caption{Performance comparison when different components employed in base model. ``R.'', ``C.'', and ``M.'' refer to the Routing mechanism, Counterfactual training, and Mixture-of-Experts, respectively. We introduce an additional Router component to MID (MRID) to specifically assess the Router’s contribution. See Appendix~\ref{app:mrid} for implementation details.}
  \label{tab:ablation-components}
\end{table}

\section{Experimental Setup}
In this section, we outline the experimental framework used to evaluate our proposed methods. Comprehensive details regarding datasets, evaluation metrics, implementation specifics, and all comparative methods are provided in App.~\ref{app:experimental_details}.
\subsection{Datasets and Implementation Details}
We evaluated our methods on two representative multimodal tasks: sarcasm detection using MMSD2.0~\citep{qin-etal-2023-mmsd2}, and sentiment analysis using MVSA-Multi~\citep{MVSA}. Standard metrics Accuracy, Precision, Recall, and F1 were utilized for sarcasm detection, while Accuracy, Macro-F1, and Weighted-F1 were used for sentiment analysis. Our experiments employed the large models Qwen2-VL-7B~\citep{wang2024qwen2} and InternVL2.5-4B~\citep{chen2025expandingperformanceboundariesopensource}, fine-tuned using LoRA~\citep{hu2022lora}. Our router model was based on CLIP~\citep{pmlr-v139-radford21a}. Inference-time debiasing hyperparameters were tuned via Bayesian optimization~\citep{NIPS2012_05311655}.

\subsection{Comparing Methods}
We compared against task-specific multimodal baselines:
Sarcasm Detection: \textbf{HFM}, \textbf{Attn-BERT}, \textbf{CMGCN}, \textbf{HKE}, \textbf{Multi-view CLIP}. Sentiment Analysis: \textbf{MVAN}, \textbf{MGNNS}, \textbf{CLMLF}, \textbf{MDSE}.
and multimodal causal debiasing methods: \textbf{TFCD}, \textbf{MCIS}, \textbf{CF-MSA}. Detailed descriptions and selection reason are in Appendix~\ref{appendix:comparing_methods}.
\section{Results and Analysis}
\begin{figure*}[!ht]
    \centering
    \includegraphics[width=0.9\linewidth]{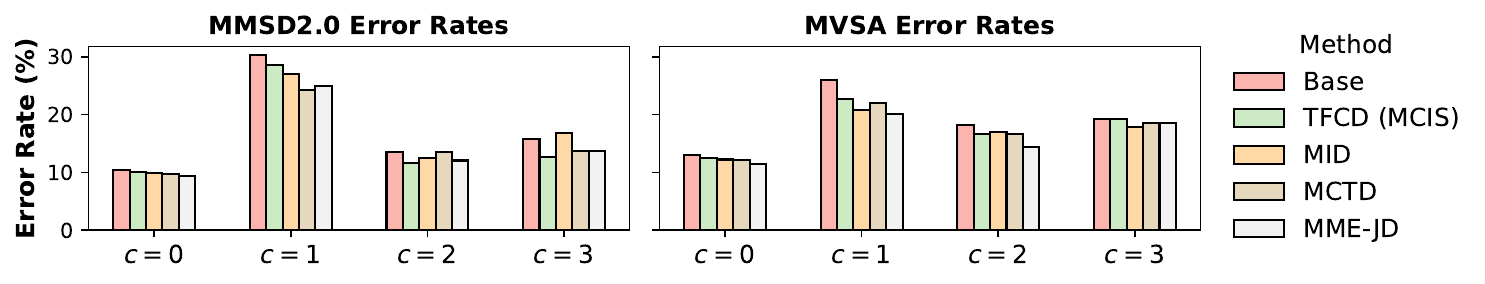}
    \caption{Comparison of error rates (\%) across different debiasing methods on MMSD2.0 and MVSA datasets regarding to debiasing category. 0/1/2/3 refer to no/image/text/both debias need.}
    \label{fig:error-rate}
\end{figure*}
\subsection{Main Results}
Tab.~\ref{tab:mmsd} and \ref{tab:mvsa} present our proposed methods’ performance comparisons on two datasets against strong MLLM baselines (InternVL2.5, Qwen2-VL) and existing task-specific methods. While existing text-based counterfactual debiasing (e.g., TFCD, MCIS) offers slight improvements, our proposed MID, which extends counterfactual inference to the multimodal level, further enhances performance. Integrating counterfactual samples during training (MCTD) also yields gains, though not consistently surpassing MID, suggesting the continued necessity for inference-time debiasing.

Crucially, our comprehensive MME-JD model, which combines counterfactual training with a router and Mixture-of-Experts, consistently delivered the most substantial improvements. It achieved top F1 scores of 85.52\% (InternVL2.5) and 87.20\% (Qwen2-VL) on MMSD2.0. These results confirm MME-JD's effectiveness in mitigating multimodal spurious correlations through the synergy of training-time strategies, expert routing, and inference-time debiasing.

\subsection{Ablation Study}
Ablation studies (Tab.~\ref{tab:ablation-components}) demonstrate the individual and combined contributions of our framework's components. For inference-time debiasing, the Router (MRID) significantly enhanced performance over MID. For training debiasing, counterfactual training (MCTD) alone improved F1 scores (e.g., +1.4 on MMSD2.0 over baseline), and the complete MME-JD model, integrating the Router, Counterfactual Training, and Mixture-of-Experts (MoE), achieved the highest F1 scores (87.20 on MMSD2.0, 62.42 on MVSA-Multi). This underscores that the deep integration of all components is crucial for optimal performance and learning robust, debiased representations.

Further analysis (Tab.~\ref{tab:ablation-modality}) on modality-specific debiasing shows that while individual image or text debiasing provides modest gains, neither matches the performance of the fully integrated MME-JD, which highlights the necessity of simultaneously addressing biases across both modalities for optimal multimodal understanding.

\subsection{Analysis on Error Rate on Categories}
To further understand the impact of our methods, we analyzed error rates across sample categories with distinct bias types following~\ref{sec:counterfactual-sample-categorize}. Results shown in Fig.~\ref{fig:error-rate} indicate that samples requiring any form of debiasing generally presented higher difficulty than those without debiasing needs ($c=0$). Among these, text-only biased samples ($c=2$) consistently showed better debiasing effectiveness compared to image-only biased samples ($c=1$). TFCD exhibited notable limitations when addressing image-biased samples ($c=1$), aligning with our expectations given its text-focused design. MID and MCTD each demonstrated advantages in different categories, underscoring the importance of integrating both inference and training-based debiasing approaches. The proposed MME-JD method, although not always outperforming all other methods within each individual category, achieved the best overall performance by effectively integrating multiple debiasing strategies.
\begin{table}[!ht]
\centering
\small
\begin{tabularx}{\linewidth}{@{\extracolsep{\fill}}l*{4}{c}@{}}
\toprule
\textbf{Methods} & \textbf{Acc} & \textbf{Prec} & \textbf{Recall} & \textbf{F1} \\ \midrule
Qwen2VL & \textbf{86.74} & \textbf{81.14} & \textbf{90.45} & \textbf{85.54} \\
\; + IDE & 87.13 & 81.2 & 91.13 & 85.89 \\
\; + TDE & 87.05 & 81.22 & 90.93 & 85.8 \\
\; + MME-JD & \textbf{88.42} & \textbf{83.19} & \textbf{91.61} & \textbf{87.2} \\ \midrule
InternVL2.5 & \textbf{85.76} & \textbf{79.91} & \textbf{89.39} & \textbf{84.38} \\
\; + IDE & 86.17 & 80.26 & 90.15 & 84.92 \\
\; + TDE & 86.21 & 80.36 & 89.97 & 84.89 \\
\; + MME-JD & \textbf{86.84} & \textbf{81.25} & \textbf{90.26} & \textbf{85.52} \\ \bottomrule
\end{tabularx}
\caption{Ablation study on the effectiveness of image and text debiasing experts within the proposed MME-JD framework.}
\label{tab:ablation-modality}
\end{table}
\subsection{Analysis on Router}
\begin{table}[t]
\centering
\small
\begin{tabular}{@{}cccccc@{}}
\toprule
\multicolumn{1}{l}{\textbf{Method}} & \textbf{Router} & \textbf{Acc.} & \textbf{Prec.} & \textbf{Recall} & \textbf{F1} \\ \midrule
Qwen2VL & trained & 88.42 & 83.19 & 91.61 & 87.20 \\
+ MME-JD & \textit{oracle} & \textit{92.40} & \textit{87.46} & \textit{96.14} & \textit{91.59} \\ \midrule
InternVL2.5 & trained & 86.84 & 81.25 & 90.26 & 85.52 \\
+ MME-JD & \textit{oracle} & \textit{89.72} & \textit{83.93} & \textit{93.45} & \textit{88.43} \\ \bottomrule
\end{tabular}
\caption{Comparison of model performance using a trained router versus an oracle router in MME-JD.}
\label{tab:router-effect}
\end{table}

\begin{table}[t]
\centering
\small
\begin{tabularx}{\linewidth}{lCCCC}
\toprule
\textbf{Model} & \textbf{Cate.} & \textbf{Prec.} & \textbf{Recall} & \textbf{F-0.5} \\
\midrule
\multirow{3}{*}{\makecell[l]{Router for\\Qwen2VL}}
  & $c{=}0$ & 77.55 & 96.40  & 80.7 \\
  & $c{=}1$ & 45.54 & 13.46 & 30.83 \\
  & $c{=}2$ & 32.00 & 12.72 & 18.20 \\
\midrule
\multirow{3}{*}{\makecell[l]{Router for\\InternVL2.5}}
  & $c{=}0$ & 81.36 & 77.12 & 80.48 \\
  & $c{=}1$ & 27.86 & 13.04 & 22.71 \\
  & $c{=}2$ & 29.82 & 42.89 & 31.75 \\
\bottomrule
\end{tabularx}
\caption{Router classification performance for different debiasing types. $c$ indicates debiasing category: 0 (No Debias), 1 (Image Debias), and 2 (Text Debias).}
\label{tab:mmd_router_metric}
\end{table}
The router component dynamically assigns input samples to suitable experts within our MME-JD framework. To evaluate its effectiveness, we compared a trained router against an oracle router (Tab.~\ref{tab:router-effect}). The oracle router, serving as an ideal upper bound (following Sec.~\ref{sec:counterfactual-sample-categorize}), demonstrated notably superior performance, highlighting both the substantial potential of the MME-JD expert architecture and the trained router’s accuracy as a key performance bottleneck.

Tab.~\ref{tab:mmd_router_metric} further shows that while the trained router effectively identifies samples not requiring debiasing ($c{=}0$), its performance degrades sharply for image ($c{=}1$) and text ($c{=}2$) debiasing cases. We attribute this weakness primarily to severe training-data imbalance—only a small minority of samples require specialized debiasing—so errors on these critical cases disproportionately limit overall MME-JD effectiveness.

To diagnose the failure modes more precisely, we analyze the class-wise confusion matrix in Tab.~\ref{tab:router-confusion}. The router exhibits a strong conservative bias toward \emph{No Debias}: 92.7\% of all predictions fall into this branch (2233/2409), yielding high recall for \emph{No Debias} (94.2\%) but low recall for \emph{Image}/\emph{Text}/\emph{Both} (6.5\%/13.3\%/7.6\%). Among all non-\emph{No Debias} ground-truth instances, 97.5\% of the errors are conservative \emph{No Debias} predictions (499 out of 512), indicating that the router abstains when uncertain rather than confusing debiasing types. Indeed, after excluding the \emph{No Debias} branch, the largest entry in each remaining row lies on the diagonal (10/36/11 for \emph{Image}/\emph{Text}/\emph{Both}), implying limited cross-type confusion once the router commits to debiasing. Class-precision by prediction is 77.7\% (\emph{No}), 22.2\% (\emph{Image}), 31.9\% (\emph{Text}), and 61.1\% (\emph{Both}). Together with the distribution statistics (Appendix, Tab.~\ref{tab:debias-dist-appendix}), these results suggest that threshold calibration or cost-sensitive training (e.g., class-balanced losses or utility-aware decision thresholds) could trade additional recall on debiasing branches for a modest precision drop, improving end-to-end routing without substantially increasing cross-class swaps, which drives our future work.

\subsection*{Qualitative evidence}
We provide two representative sarcasm–detection case studies in Appendix~\ref{app:case_study}. Both examples apply our masking-based debiasing to discount background-only evidence while preserving task-relevant semantics.

\begin{table}[t]
\centering
\small
\begin{tabular}{lcccc}
\toprule
\textbf{True} $\backslash$ \textbf{Pred} & \textbf{No Debias} & \textbf{Image} & \textbf{Text} & \textbf{Both} \\
\midrule
No Debias      & 1{,}734 & 29 & 72 & 5 \\
Image Debias   & 140  & 10 & 3  & 1 \\
Text Debias    & 230  & 3  & 36 & 1 \\
Double Debias  & 129  & 3  & 2  & 11 \\
\bottomrule
\end{tabular}
\caption{Confusion matrix of the trained router.}
\label{tab:router-confusion}
\end{table}

\section{Conclusion}
In this paper, we addressed the critical issue of spurious multimodal biases that impair the robustness and generalization of MLLMs. Prior approaches lacked comprehensive multimodal analysis and training-time adaptability. We first developed a fine-grained multimodal causal framework, explicitly distinguishing spurious context from semantic content, thereby extending previous inference-time adjustments to multimodal scenarios. Building upon this, we introduced MME-JD, incorporating counterfactual content into training and employing a Mixture-of-Experts architecture with dynamic routing for adaptive, sample-specific debiasing. Extensive experiments demonstrate that MME-JD significantly surpasses simpler debiasing methods and existing benchmarks. Future research will explore enhancing counterfactual generation techniques and refining router designs for enhancement.
\section*{Acknowledgments}
This work is supported by the National Natural Science Foundation of China (62076008).

\section*{Limitation}
Despite the improvements demonstrated, our approach has several limitations that open avenues for future research:

\paragraph{Accuracy of Spurious Context Identification:}
The efficacy of our framework critically depends on the precise identification of spurious contexts versus core semantic content. Our current methodology, employing prompt engineering and attention-based mechanisms for semantic extraction, faces inherent challenges. Prompt engineering can inadvertently introduce new biases, while attention scores may not always reliably pinpoint genuinely spurious regions. Therefore, despite the advanced nature of LLMs, inaccuracies in this initial extraction phase can propagate the effectiveness of subsequent debiasing efforts. Future studies should integrate additional robust and objective methodologies alongside these advanced models to further enhance extraction accuracy and reliability.

\paragraph{Router Mechanism Complexity and Stability:}
While the router mechanism in MME-JD is designed to manage the variability of counterfactual sample quality, its own training and calibration require careful attention and can be intricate. This complexity might risk it becoming a performance bottleneck or introducing sensitivities. For instance, ensuring the router generalizes well and avoids ‘dependency loops’ (where its performance is overly reliant on specific states of other components it's trying to manage) is crucial. Developing more inherently robust router architectures, along with streamlined and stable training strategies, presents a key direction for future improvement.

In addition, our router faces severe class imbalance, since only around 10\% of MMSD2.0 samples require explicit debiasing, as detailed in Table~\ref{tab:debias-dist-appendix}. We did not apply resampling, loss weighting, or focal loss during router training, prioritizing precision over recall. This conservative choice avoids mismatched debiasing but leads to weaker performance on minority categories (e.g., image-only debias). Exploring class-aware or curriculum-based router training strategies is an important direction for future work.

\paragraph{Linearity Assumption in Causal Mediation:}
Our current causal mediation framework approximates bias removal using linear relationships. However, the interactions between biases and core semantics within complex Multimodal Large Language Models (MLLMs) are likely to be highly nonlinear. This linearity assumption may therefore limit our model's capacity to fully neutralize biases, particularly in scenarios requiring simultaneous debiasing of both textual and visual modalities (e.g., in Fig~\ref{fig:error-rate}, MME-JD did not perform the best, and sometimes text-specific methods (TFCD) could surpass multimodal methods). A significant avenue for future research lies in investigating and integrating nonlinear debiasing strategies to more comprehensively model and mitigate these intricate multimodal interactions, thereby further enhancing model robustness.

\paragraph{Inference Efficiency of MME-JD}
The MME-JD framework's adaptive inference process, while effective for debiasing, can impact its overall efficiency. Key steps, including the on-the-fly generation of counterfactual inputs $(\hat{t}, \hat{i}) $, router processing, and the subsequent execution of selected expert(s), collectively contribute to increased latency and computational resource demands compared to simpler methods. While we currently employ independent models for each expert, we have tried MoE-LoRA~\citep{luo2024moeloracontrastivelearningguided} to reduce the training and inference cost, but it did not yield comparable debiasing performance. How to develop more successful parameter-efficient expert architectures that maintain high debiasing capabilities is worth considering in the future.

\section*{Ethical Considerations}
\paragraph{Potential Risks}
While our methods effectively reduce known biases, they depend on identifying spurious correlations accurately. Errors or oversights in distinguishing between semantic and spurious contexts could inadvertently reinforce existing biases or introduce new ones. Practitioners must be cautious and continually validate debiasing processes to prevent unintended consequences.

\paragraph{Use of AI Assistants}
We have employed ChatGPT as a writing assistant, primarily for polishing the text after the initial composition.
\nocite{wang2024comprehensivereviewmultimodallarge}
\nocite{hosseini2025seeingwhatstherespurious}
\nocite{Niu_2021_CVPR}
\nocite{Kim_2023_CVPR}
\nocite{ye2024mmspubenchbetterunderstandingspurious}
\nocite{Agarwal_2020_CVPR}
\nocite{Chen_2020_CVPR}
\nocite{Howard_2024_CVPR}
\nocite{NEURIPS2024_95c6ae3f}
\nocite{pmlr-v235-zhang24as}
\nocite{Marani_2024}
\nocite{8951256}
\nocite{ijcai2024p0887}
\nocite{SOLEYMANI20173}
\nocite{info13080399}
\nocite{pmlr-v139-radford21a}
\nocite{bai2023qwen}
\nocite{wang2024qwen2}
\nocite{Chen_2024_CVPR}
\nocite{chen2024far}
\nocite{chen2025expandingperformanceboundariesopensource}
\nocite{Agrawal_2018_CVPR}
\nocite{ijcai2024p739}
\nocite{Yang_2024_CVPR}
\nocite{chen2024multimodalsentimentanalysisbased}
\nocite{9246699}
\nocite{10445820}
\nocite{10.1145/3501714.3501734}
\nocite{10.1007/978-3-031-73636-0_27}
\nocite{NIPS2012_05311655}
\nocite{MVSA}
\nocite{hu2022lora}
\nocite{rajbhandari2020zeromemoryoptimizationstraining}
\nocite{bai2025hallucinationmultimodallargelanguage}
\nocite{yu2025aligningmultimodalllmhuman}
\nocite{wu2024causalitylargelanguagemodels}
\nocite{zhang2024debiasingmultimodallargelanguage}
\nocite{openai2024gpt4technicalreport}
\nocite{NEURIPS2023_6dcf277e}
\nocite{luo2024moeloracontrastivelearningguided}
\nocite{patil-etal-2023-debiasing}
\nocite{qin-etal-2023-mmsd2}
\nocite{weng-etal-2024-images}
\nocite{he-etal-2022-cpl}
\nocite{castro-etal-2019-towards}
\nocite{chen-etal-2024-quantifying}
\nocite{liu-etal-2024-ce}
\nocite{cai-etal-2019-multi}
\nocite{pan-etal-2020-modeling}
\nocite{liu-etal-2022-towards-multi-modal}
\nocite{liang-etal-2022-multi}
\nocite{yang-etal-2021-multimodal}
\nocite{li-etal-2022-clmlf}
\nocite{feder-etal-2022-causal}
\nocite{wang-etal-2021-counterfactual-adversarial}
\nocite{wang-etal-2024-multimodal}
\nocite{zheng-etal-2024-llamafactory}
\nocite{wolf-etal-2020-transformers}
\bibliography{anthology,custom}

\begin{thebibliography}{69}
\providecommand{\natexlab}[1]{#1}

\bibitem[{Agarwal et~al.(2020)Agarwal, Shetty, and Fritz}]{Agarwal_2020_CVPR}
Vedika Agarwal, Rakshith Shetty, and Mario Fritz. 2020.
\newblock Towards causal vqa: Revealing and reducing spurious correlations by invariant and covariant semantic editing.
\newblock In \emph{Proceedings of the IEEE/CVF Conference on Computer Vision and Pattern Recognition (CVPR)}.

\bibitem[{Agrawal et~al.(2018)Agrawal, Batra, Parikh, and Kembhavi}]{Agrawal_2018_CVPR}
Aishwarya Agrawal, Dhruv Batra, Devi Parikh, and Aniruddha Kembhavi. 2018.
\newblock Don't just assume; look and answer: Overcoming priors for visual question answering.
\newblock In \emph{Proceedings of the IEEE Conference on Computer Vision and Pattern Recognition (CVPR)}.

\bibitem[{Bai et~al.(2023)Bai, Bai, Chu, Cui, Dang, Deng, Fan, Ge, Han, Huang et~al.}]{bai2023qwen}
Jinze Bai, Shuai Bai, Yunfei Chu, Zeyu Cui, Kai Dang, Xiaodong Deng, Yang Fan, Wenbin Ge, Yu~Han, Fei Huang, and 1 others. 2023.
\newblock Qwen technical report.
\newblock \emph{arXiv preprint arXiv:2309.16609}.

\bibitem[{Bai et~al.(2025)Bai, Wang, Xiao, He, Han, Zhang, and Shou}]{bai2025hallucinationmultimodallargelanguage}
Zechen Bai, Pichao Wang, Tianjun Xiao, Tong He, Zongbo Han, Zheng Zhang, and Mike~Zheng Shou. 2025.
\newblock \href {https://arxiv.org/abs/2404.18930} {Hallucination of multimodal large language models: A survey}.
\newblock \emph{Preprint}, arXiv:2404.18930.

\bibitem[{Băroiu and Trăușan-Matu(2022)}]{info13080399}
Alexandru-Costin Băroiu and Ștefan Trăușan-Matu. 2022.
\newblock \href {https://doi.org/10.3390/info13080399} {Automatic sarcasm detection: Systematic literature review}.
\newblock \emph{Information}, 13(8).

\bibitem[{Cai et~al.(2019)Cai, Cai, and Wan}]{cai-etal-2019-multi}
Yitao Cai, Huiyu Cai, and Xiaojun Wan. 2019.
\newblock \href {https://doi.org/10.18653/v1/P19-1239} {Multi-modal sarcasm detection in {T}witter with hierarchical fusion model}.
\newblock In \emph{Proceedings of the 57th Annual Meeting of the Association for Computational Linguistics}, pages 2506--2515, Florence, Italy. Association for Computational Linguistics.

\bibitem[{Castro et~al.(2019)Castro, Hazarika, P{\'e}rez-Rosas, Zimmermann, Mihalcea, and Poria}]{castro-etal-2019-towards}
Santiago Castro, Devamanyu Hazarika, Ver{\'o}nica P{\'e}rez-Rosas, Roger Zimmermann, Rada Mihalcea, and Soujanya Poria. 2019.
\newblock \href {https://doi.org/10.18653/v1/P19-1455} {Towards multimodal sarcasm detection (an {\_}{O}bviously{\_} perfect paper)}.
\newblock In \emph{Proceedings of the 57th Annual Meeting of the Association for Computational Linguistics}, pages 4619--4629, Florence, Italy. Association for Computational Linguistics.

\bibitem[{Chakraborty et~al.(2020)Chakraborty, Bhattacharyya, and Bag}]{8951256}
Koyel Chakraborty, Siddhartha Bhattacharyya, and Rajib Bag. 2020.
\newblock \href {https://doi.org/10.1109/TCSS.2019.2956957} {A survey of sentiment analysis from social media data}.
\newblock \emph{IEEE Transactions on Computational Social Systems}, 7(2):450--464.

\bibitem[{Chen et~al.(2024{\natexlab{a}})Chen, Huang, Ge, Huang, and Bao}]{chen2024multimodalsentimentanalysisbased}
Fuhai Chen, Pengpeng Huang, Xuri Ge, Jie Huang, and Zishuo Bao. 2024{\natexlab{a}}.
\newblock \href {https://arxiv.org/abs/2412.07292} {Multimodal sentiment analysis based on causal reasoning}.
\newblock \emph{Preprint}, arXiv:2412.07292.

\bibitem[{Chen et~al.(2020)Chen, Yan, Xiao, Zhang, Pu, and Zhuang}]{Chen_2020_CVPR}
Long Chen, Xin Yan, Jun Xiao, Hanwang Zhang, Shiliang Pu, and Yueting Zhuang. 2020.
\newblock Counterfactual samples synthesizing for robust visual question answering.
\newblock In \emph{Proceedings of the IEEE/CVF Conference on Computer Vision and Pattern Recognition (CVPR)}.

\bibitem[{Chen et~al.(2024{\natexlab{b}})Chen, Cao, Zhang, and Lu}]{chen-etal-2024-quantifying}
Meiqi Chen, Yixin Cao, Yan Zhang, and Chaochao Lu. 2024{\natexlab{b}}.
\newblock \href {https://doi.org/10.18653/v1/2024.findings-emnlp.960} {Quantifying and mitigating unimodal biases in multimodal large language models: A causal perspective}.
\newblock In \emph{Findings of the Association for Computational Linguistics: EMNLP 2024}, pages 16449--16469, Miami, Florida, USA. Association for Computational Linguistics.

\bibitem[{Chen et~al.(2025)Chen, Wang, Cao, Liu, Gao, Cui, Zhu, Ye, Tian, Liu, Gu, Wang, Li, Ren, Chen, Luo, Wang, Jiang, Wang, He, Shi, Zhang, Lv, Wang, Shao, Chu, Tu, He, Wu, Deng, Ge, Chen, Zhang, Wang, Dou, Lu, Zhu, Lu, Lin, Qiao, Dai, and Wang}]{chen2025expandingperformanceboundariesopensource}
Zhe Chen, Weiyun Wang, Yue Cao, Yangzhou Liu, Zhangwei Gao, Erfei Cui, Jinguo Zhu, Shenglong Ye, Hao Tian, Zhaoyang Liu, Lixin Gu, Xuehui Wang, Qingyun Li, Yimin Ren, Zixuan Chen, Jiapeng Luo, Jiahao Wang, Tan Jiang, Bo~Wang, and 23 others. 2025.
\newblock \href {https://arxiv.org/abs/2412.05271} {Expanding performance boundaries of open-source multimodal models with model, data, and test-time scaling}.
\newblock \emph{Preprint}, arXiv:2412.05271.

\bibitem[{Chen et~al.(2024{\natexlab{c}})Chen, Wang, Tian, Ye, Gao, Cui, Tong, Hu, Luo, Ma et~al.}]{chen2024far}
Zhe Chen, Weiyun Wang, Hao Tian, Shenglong Ye, Zhangwei Gao, Erfei Cui, Wenwen Tong, Kongzhi Hu, Jiapeng Luo, Zheng Ma, and 1 others. 2024{\natexlab{c}}.
\newblock How far are we to gpt-4v? closing the gap to commercial multimodal models with open-source suites.
\newblock \emph{Science China Information Sciences}, 67(12):220101.

\bibitem[{Chen et~al.(2024{\natexlab{d}})Chen, Wu, Wang, Su, Chen, Xing, Zhong, Zhang, Zhu, Lu, Li, Luo, Lu, Qiao, and Dai}]{Chen_2024_CVPR}
Zhe Chen, Jiannan Wu, Wenhai Wang, Weijie Su, Guo Chen, Sen Xing, Muyan Zhong, Qinglong Zhang, Xizhou Zhu, Lewei Lu, Bin Li, Ping Luo, Tong Lu, Yu~Qiao, and Jifeng Dai. 2024{\natexlab{d}}.
\newblock Internvl: Scaling up vision foundation models and aligning for generic visual-linguistic tasks.
\newblock In \emph{Proceedings of the IEEE/CVF Conference on Computer Vision and Pattern Recognition (CVPR)}, pages 24185--24198.

\bibitem[{Farabi et~al.(2024)Farabi, Ranasinghe, Kanojia, Kong, and Zampieri}]{ijcai2024p0887}
Shafkat Farabi, Tharindu Ranasinghe, Diptesh Kanojia, Yu~Kong, and Marcos Zampieri. 2024.
\newblock \href {https://doi.org/10.24963/ijcai.2024/887} {A survey of multimodal sarcasm detection}.
\newblock In \emph{Proceedings of the Thirty-Third International Joint Conference on Artificial Intelligence, {IJCAI-24}}, pages 8020--8028. International Joint Conferences on Artificial Intelligence Organization.
\newblock Survey Track.

\bibitem[{Feder et~al.(2022)Feder, Keith, Manzoor, Pryzant, Sridhar, Wood-Doughty, Eisenstein, Grimmer, Reichart, Roberts, Stewart, Veitch, and Yang}]{feder-etal-2022-causal}
Amir Feder, Katherine~A. Keith, Emaad Manzoor, Reid Pryzant, Dhanya Sridhar, Zach Wood-Doughty, Jacob Eisenstein, Justin Grimmer, Roi Reichart, Margaret~E. Roberts, Brandon~M. Stewart, Victor Veitch, and Diyi Yang. 2022.
\newblock \href {https://doi.org/10.1162/tacl_a_00511} {Causal inference in natural language processing: Estimation, prediction, interpretation and beyond}.
\newblock \emph{Transactions of the Association for Computational Linguistics}, 10:1138--1158.

\bibitem[{Golovanevsky et~al.(2025)Golovanevsky, Rudman, Palit, Singh, and Eickhoff}]{golovanevsky2024notice}
Michal Golovanevsky, William Rudman, Vedant Palit, Ritambhara Singh, and Carsten Eickhoff. 2025.
\newblock \href {https://arxiv.org/abs/2406.16320} {What do vlms notice? a mechanistic interpretability pipeline for gaussian-noise-free text-image corruption and evaluation}.
\newblock \emph{Preprint}, arXiv:2406.16320.

\bibitem[{He et~al.(2022)He, Yang, Feng, Fu, Akula, Jampani, Narayana, Basu, Wang, and Wang}]{he-etal-2022-cpl}
Xuehai He, Diji Yang, Weixi Feng, Tsu-Jui Fu, Arjun Akula, Varun Jampani, Pradyumna Narayana, Sugato Basu, William~Yang Wang, and Xin Wang. 2022.
\newblock \href {https://doi.org/10.18653/v1/2022.emnlp-main.224} {{CPL}: Counterfactual prompt learning for vision and language models}.
\newblock In \emph{Proceedings of the 2022 Conference on Empirical Methods in Natural Language Processing}, pages 3407--3418, Abu Dhabi, United Arab Emirates. Association for Computational Linguistics.

\bibitem[{Hosseini et~al.(2025)Hosseini, Nawathe, Moayeri, Balasubramanian, and Feizi}]{hosseini2025seeingwhatstherespurious}
Parsa Hosseini, Sumit Nawathe, Mazda Moayeri, Sriram Balasubramanian, and Soheil Feizi. 2025.
\newblock \href {https://arxiv.org/abs/2503.08884} {Seeing what's not there: Spurious correlation in multimodal llms}.
\newblock \emph{Preprint}, arXiv:2503.08884.

\bibitem[{Howard et~al.(2024)Howard, Madasu, Le, Moreno, Bhiwandiwalla, and Lal}]{Howard_2024_CVPR}
Phillip Howard, Avinash Madasu, Tiep Le, Gustavo~Lujan Moreno, Anahita Bhiwandiwalla, and Vasudev Lal. 2024.
\newblock Socialcounterfactuals: Probing and mitigating intersectional social biases in vision-language models with counterfactual examples.
\newblock In \emph{Proceedings of the IEEE/CVF Conference on Computer Vision and Pattern Recognition (CVPR)}, pages 11975--11985.

\bibitem[{Hu et~al.(2022)Hu, yelong shen, Wallis, Allen-Zhu, Li, Wang, Wang, and Chen}]{hu2022lora}
Edward~J Hu, yelong shen, Phillip Wallis, Zeyuan Allen-Zhu, Yuanzhi Li, Shean Wang, Lu~Wang, and Weizhu Chen. 2022.
\newblock \href {https://openreview.net/forum?id=nZeVKeeFYf9} {Lo{RA}: Low-rank adaptation of large language models}.
\newblock In \emph{International Conference on Learning Representations}.

\bibitem[{Kim et~al.(2023)Kim, Koepke, Schmid, and Akata}]{Kim_2023_CVPR}
Jae~Myung Kim, A.~Sophia Koepke, Cordelia Schmid, and Zeynep Akata. 2023.
\newblock Exposing and mitigating spurious correlations for cross-modal retrieval.
\newblock In \emph{Proceedings of the IEEE/CVF Conference on Computer Vision and Pattern Recognition (CVPR) Workshops}, pages 2585--2595.

\bibitem[{Le et~al.(2023)Le, LAL, and Howard}]{le2023coco}
Tiep Le, VASUDEV LAL, and Phillip Howard. 2023.
\newblock \href {https://proceedings.neurips.cc/paper_files/paper/2023/file/e14e4cb8266184ceb234973dfe07faed-Paper-Datasets_and_Benchmarks.pdf} {Coco-counterfactuals: Automatically constructed counterfactual examples for image-text pairs}.
\newblock In \emph{Advances in Neural Information Processing Systems}, volume~36, pages 71195--71221. Curran Associates, Inc.

\bibitem[{Leng et~al.(2024)Leng, Zhang, Chen, Li, Lu, Miao, and Bing}]{leng2024vcd}
Sicong Leng, Hang Zhang, Guanzheng Chen, Xin Li, Shijian Lu, Chunyan Miao, and Lidong Bing. 2024.
\newblock Mitigating object hallucinations in large vision-language models through visual contrastive decoding.
\newblock In \emph{Proceedings of the IEEE/CVF Conference on Computer Vision and Pattern Recognition (CVPR)}, pages 13872--13882.

\bibitem[{Li et~al.(2024)Li, Wang, Luo, Wu, and Jiang}]{10445820}
Jingzhe Li, Chengji Wang, Zhiming Luo, Yuxian Wu, and Xingpeng Jiang. 2024.
\newblock \href {https://doi.org/10.1109/ICASSP48485.2024.10445820} {Modality-dependent sentiments exploring for multi-modal sentiment classification}.
\newblock In \emph{ICASSP 2024 - 2024 IEEE International Conference on Acoustics, Speech and Signal Processing (ICASSP)}, pages 7930--7934.

\bibitem[{Li et~al.(2022)Li, Xu, Zhu, and Zhao}]{li-etal-2022-clmlf}
Zhen Li, Bing Xu, Conghui Zhu, and Tiejun Zhao. 2022.
\newblock \href {https://doi.org/10.18653/v1/2022.findings-naacl.175} {{CLMLF}:a contrastive learning and multi-layer fusion method for multimodal sentiment detection}.
\newblock In \emph{Findings of the Association for Computational Linguistics: NAACL 2022}, pages 2282--2294, Seattle, United States. Association for Computational Linguistics.

\bibitem[{Liang et~al.(2022)Liang, Lou, Li, Yang, Gui, He, Pei, and Xu}]{liang-etal-2022-multi}
Bin Liang, Chenwei Lou, Xiang Li, Min Yang, Lin Gui, Yulan He, Wenjie Pei, and Ruifeng Xu. 2022.
\newblock \href {https://doi.org/10.18653/v1/2022.acl-long.124} {Multi-modal sarcasm detection via cross-modal graph convolutional network}.
\newblock In \emph{Proceedings of the 60th Annual Meeting of the Association for Computational Linguistics (Volume 1: Long Papers)}, pages 1767--1777, Dublin, Ireland. Association for Computational Linguistics.

\bibitem[{Liu et~al.(2023)Liu, Li, Wu, and Lee}]{NEURIPS2023_6dcf277e}
Haotian Liu, Chunyuan Li, Qingyang Wu, and Yong~Jae Lee. 2023.
\newblock \href {https://proceedings.neurips.cc/paper_files/paper/2023/file/6dcf277ea32ce3288914faf369fe6de0-Paper-Conference.pdf} {Visual instruction tuning}.
\newblock In \emph{Advances in Neural Information Processing Systems}, volume~36, pages 34892--34916. Curran Associates, Inc.

\bibitem[{Liu et~al.(2024)Liu, Wang, Zhu, Wang, and Wang}]{liu-etal-2024-ce}
Hongcheng Liu, Pingjie Wang, Zhiyuan Zhu, Yanfeng Wang, and Yu~Wang. 2024.
\newblock \href {https://aclanthology.org/2024.lrec-main.264/} {{CE}-{VDG}: Counterfactual entropy-based bias reduction for video-grounded dialogue generation}.
\newblock In \emph{Proceedings of the 2024 Joint International Conference on Computational Linguistics, Language Resources and Evaluation (LREC-COLING 2024)}, pages 2958--2968, Torino, Italia. ELRA and ICCL.

\bibitem[{Liu et~al.(2022)Liu, Wang, and Li}]{liu-etal-2022-towards-multi-modal}
Hui Liu, Wenya Wang, and Haoliang Li. 2022.
\newblock \href {https://doi.org/10.18653/v1/2022.emnlp-main.333} {Towards multi-modal sarcasm detection via hierarchical congruity modeling with knowledge enhancement}.
\newblock In \emph{Proceedings of the 2022 Conference on Empirical Methods in Natural Language Processing}, pages 4995--5006, Abu Dhabi, United Arab Emirates. Association for Computational Linguistics.

\bibitem[{Luo et~al.(2024)Luo, Lei, Lei, Liu, He, Zhao, and Liu}]{luo2024moeloracontrastivelearningguided}
Tongxu Luo, Jiahe Lei, Fangyu Lei, Weihao Liu, Shizhu He, Jun Zhao, and Kang Liu. 2024.
\newblock \href {https://arxiv.org/abs/2402.12851} {Moelora: Contrastive learning guided mixture of experts on parameter-efficient fine-tuning for large language models}.
\newblock \emph{Preprint}, arXiv:2402.12851.

\bibitem[{Marani et~al.(2024)Marani, Hanini, Malayarukil, Christodoulidis, Vakalopoulou, and Ferrante}]{Marani_2024}
Badr-Eddine Marani, Mohamed Hanini, Nihitha Malayarukil, Stergios Christodoulidis, Maria Vakalopoulou, and Enzo Ferrante. 2024.
\newblock \href {https://doi.org/10.1007/978-3-031-73202-7_24} {\emph{ViG-Bias: Visually Grounded Bias Discovery and Mitigation}}, page 414–429.
\newblock Springer Nature Switzerland.

\bibitem[{Niu et~al.(2016)Niu, Zhu, Pang, and El{-}Saddik}]{MVSA}
Teng Niu, Shiai Zhu, Lei Pang, and Abdulmotaleb El{-}Saddik. 2016.
\newblock Sentiment analysis on multi-view social data.
\newblock In \emph{MultiMedia Modeling}, page 15–27.

\bibitem[{Niu et~al.(2021)Niu, Tang, Zhang, Lu, Hua, and Wen}]{Niu_2021_CVPR}
Yulei Niu, Kaihua Tang, Hanwang Zhang, Zhiwu Lu, Xian-Sheng Hua, and Ji-Rong Wen. 2021.
\newblock Counterfactual vqa: A cause-effect look at language bias.
\newblock In \emph{Proceedings of the IEEE/CVF Conference on Computer Vision and Pattern Recognition (CVPR)}, pages 12700--12710.

\bibitem[{OpenAI et~al.(2024)OpenAI, Achiam, Adler, Agarwal, Ahmad, Akkaya, Aleman, Almeida, Altenschmidt, Altman, Anadkat, Avila, Babuschkin, Balaji, Balcom, Baltescu, Bao, Bavarian, Belgum, Bello, Berdine, Bernadett-Shapiro, Berner, Bogdonoff, Boiko, Boyd, Brakman, Brockman, Brooks, Brundage, Button, Cai, Campbell, Cann, Carey, Carlson, Carmichael, Chan, Chang, Chantzis, Chen, Chen, Chen, Chen, Chen, Chess, Cho, Chu, Chung, Cummings, Currier, Dai, Decareaux, Degry, Deutsch, Deville, Dhar, Dohan, Dowling, Dunning, Ecoffet, Eleti, Eloundou, Farhi, Fedus, Felix, Fishman, Forte, Fulford, Gao, Georges, Gibson, Goel, Gogineni, Goh, Gontijo-Lopes, Gordon, Grafstein, Gray, Greene, Gross, Gu, Guo, Hallacy, Han, Harris, He, Heaton, Heidecke, Hesse, Hickey, Hickey, Hoeschele, Houghton, Hsu, Hu, Hu, Huizinga, Jain, Jain, Jang, Jiang, Jiang, Jin, Jin, Jomoto, Jonn, Jun, Kaftan, Łukasz Kaiser, Kamali, Kanitscheider, Keskar, Khan, Kilpatrick, Kim, Kim, Kim, Kirchner, Kiros, Knight, Kokotajlo, Łukasz Kondraciuk,
  Kondrich, Konstantinidis, Kosic, Krueger, Kuo, Lampe, Lan, Lee, Leike, Leung, Levy, Li, Lim, Lin, Lin, Litwin, Lopez, Lowe, Lue, Makanju, Malfacini, Manning, Markov, Markovski, Martin, Mayer, Mayne, McGrew, McKinney, McLeavey, McMillan, McNeil, Medina, Mehta, Menick, Metz, Mishchenko, Mishkin, Monaco, Morikawa, Mossing, Mu, Murati, Murk, Mély, Nair, Nakano, Nayak, Neelakantan, Ngo, Noh, Ouyang, O'Keefe, Pachocki, Paino, Palermo, Pantuliano, Parascandolo, Parish, Parparita, Passos, Pavlov, Peng, Perelman, de~Avila Belbute~Peres, Petrov, de~Oliveira~Pinto, Michael, Pokorny, Pokrass, Pong, Powell, Power, Power, Proehl, Puri, Radford, Rae, Ramesh, Raymond, Real, Rimbach, Ross, Rotsted, Roussez, Ryder, Saltarelli, Sanders, Santurkar, Sastry, Schmidt, Schnurr, Schulman, Selsam, Sheppard, Sherbakov, Shieh, Shoker, Shyam, Sidor, Sigler, Simens, Sitkin, Slama, Sohl, Sokolowsky, Song, Staudacher, Such, Summers, Sutskever, Tang, Tezak, Thompson, Tillet, Tootoonchian, Tseng, Tuggle, Turley, Tworek, Uribe, Vallone,
  Vijayvergiya, Voss, Wainwright, Wang, Wang, Wang, Ward, Wei, Weinmann, Welihinda, Welinder, Weng, Weng, Wiethoff, Willner, Winter, Wolrich, Wong, Workman, Wu, Wu, Wu, Xiao, Xu, Yoo, Yu, Yuan, Zaremba, Zellers, Zhang, Zhang, Zhao, Zheng, Zhuang, Zhuk, and Zoph}]{openai2024gpt4technicalreport}
OpenAI, Josh Achiam, Steven Adler, Sandhini Agarwal, Lama Ahmad, Ilge Akkaya, Florencia~Leoni Aleman, Diogo Almeida, Janko Altenschmidt, Sam Altman, Shyamal Anadkat, Red Avila, Igor Babuschkin, Suchir Balaji, Valerie Balcom, Paul Baltescu, Haiming Bao, Mohammad Bavarian, Jeff Belgum, and 262 others. 2024.
\newblock \href {https://arxiv.org/abs/2303.08774} {Gpt-4 technical report}.
\newblock \emph{Preprint}, arXiv:2303.08774.

\bibitem[{Palit et~al.(2023)Palit, Pandey, Arora, and Liang}]{palit2023vlm}
Vedant Palit, Rohan Pandey, Aryaman Arora, and Paul~Pu Liang. 2023.
\newblock Towards vision-language mechanistic interpretability: A causal tracing tool for blip.
\newblock In \emph{Proceedings of the IEEE/CVF International Conference on Computer Vision (ICCV) Workshops}, pages 2856--2861.

\bibitem[{Pan et~al.(2020)Pan, Lin, Fu, Qi, and Wang}]{pan-etal-2020-modeling}
Hongliang Pan, Zheng Lin, Peng Fu, Yatao Qi, and Weiping Wang. 2020.
\newblock \href {https://doi.org/10.18653/v1/2020.findings-emnlp.124} {Modeling intra and inter-modality incongruity for multi-modal sarcasm detection}.
\newblock In \emph{Findings of the Association for Computational Linguistics: EMNLP 2020}, pages 1383--1392, Online. Association for Computational Linguistics.

\bibitem[{Patil et~al.(2023)Patil, Maharana, and Bansal}]{patil-etal-2023-debiasing}
Vaidehi Patil, Adyasha Maharana, and Mohit Bansal. 2023.
\newblock \href {https://doi.org/10.18653/v1/2023.findings-emnlp.270} {Debiasing multimodal models via causal information minimization}.
\newblock In \emph{Findings of the Association for Computational Linguistics: EMNLP 2023}, pages 4108--4123, Singapore. Association for Computational Linguistics.

\bibitem[{Pearl(2022)}]{10.1145/3501714.3501734}
Judea Pearl. 2022.
\newblock \href {https://doi.org/10.1145/3501714.3501734} {\emph{Causal Diagrams for Empirical Research (With Discussions)}}, 1 edition, page 255–316.
\newblock Association for Computing Machinery, New York, NY, USA.

\bibitem[{Qin et~al.(2023)Qin, Huang, Chen, Cai, Zhang, Liang, Che, and Xu}]{qin-etal-2023-mmsd2}
Libo Qin, Shijue Huang, Qiguang Chen, Chenran Cai, Yudi Zhang, Bin Liang, Wanxiang Che, and Ruifeng Xu. 2023.
\newblock \href {https://doi.org/10.18653/v1/2023.findings-acl.689} {{MMSD}2.0: Towards a reliable multi-modal sarcasm detection system}.
\newblock In \emph{Findings of the Association for Computational Linguistics: ACL 2023}, pages 10834--10845, Toronto, Canada. Association for Computational Linguistics.

\bibitem[{Radford et~al.(2021)Radford, Kim, Hallacy, Ramesh, Goh, Agarwal, Sastry, Askell, Mishkin, Clark, Krueger, and Sutskever}]{pmlr-v139-radford21a}
Alec Radford, Jong~Wook Kim, Chris Hallacy, Aditya Ramesh, Gabriel Goh, Sandhini Agarwal, Girish Sastry, Amanda Askell, Pamela Mishkin, Jack Clark, Gretchen Krueger, and Ilya Sutskever. 2021.
\newblock \href {https://proceedings.mlr.press/v139/radford21a.html} {Learning transferable visual models from natural language supervision}.
\newblock In \emph{Proceedings of the 38th International Conference on Machine Learning}, volume 139 of \emph{Proceedings of Machine Learning Research}, pages 8748--8763. PMLR.

\bibitem[{Rajbhandari et~al.(2020)Rajbhandari, Rasley, Ruwase, and He}]{rajbhandari2020zeromemoryoptimizationstraining}
Samyam Rajbhandari, Jeff Rasley, Olatunji Ruwase, and Yuxiong He. 2020.
\newblock \href {https://arxiv.org/abs/1910.02054} {Zero: Memory optimizations toward training trillion parameter models}.
\newblock \emph{Preprint}, arXiv:1910.02054.

\bibitem[{Shekhar et~al.(2017)Shekhar, Pezzelle, Klimovich, Herbelot, Nabi, Sangineto, and Bernardi}]{foil2018}
Ravi Shekhar, Sandro Pezzelle, Yauhen Klimovich, Aurélie Herbelot, Moin Nabi, Enver Sangineto, and Raffaella Bernardi. 2017.
\newblock \href {https://doi.org/10.18653/v1/p17-1024} {Foil it! find one mismatch between image and language caption}.
\newblock In \emph{Proceedings of the 55th Annual Meeting of the Association for Computational Linguistics (Volume 1: Long Papers)}. Association for Computational Linguistics.

\bibitem[{Snoek et~al.(2012)Snoek, Larochelle, and Adams}]{NIPS2012_05311655}
Jasper Snoek, Hugo Larochelle, and Ryan~P Adams. 2012.
\newblock \href {https://proceedings.neurips.cc/paper_files/paper/2012/file/05311655a15b75fab86956663e1819cd-Paper.pdf} {Practical bayesian optimization of machine learning algorithms}.
\newblock In \emph{Advances in Neural Information Processing Systems}, volume~25. Curran Associates, Inc.

\bibitem[{Soleymani et~al.(2017)Soleymani, Garcia, Jou, Schuller, Chang, and Pantic}]{SOLEYMANI20173}
Mohammad Soleymani, David Garcia, Brendan Jou, Björn Schuller, Shih-Fu Chang, and Maja Pantic. 2017.
\newblock \href {https://doi.org/10.1016/j.imavis.2017.08.003} {A survey of multimodal sentiment analysis}.
\newblock \emph{Image and Vision Computing}, 65:3--14.
\newblock Multimodal Sentiment Analysis and Mining in the Wild Image and Vision Computing.

\bibitem[{Thrush et~al.(2022)Thrush, Jiang, Bartolo, Singh, Williams, Kiela, and Ross}]{thrush2022winoground}
Tristan Thrush, Ryan Jiang, Max Bartolo, Amanpreet Singh, Adina Williams, Douwe Kiela, and Candace Ross. 2022.
\newblock \href {https://arxiv.org/abs/2204.03162} {Winoground: Probing vision and language models for visio-linguistic compositionality}.
\newblock \emph{Preprint}, arXiv:2204.03162.

\bibitem[{Varma et~al.(2024)Varma, Delbrouck, Chen, Chaudhari, and Langlotz}]{NEURIPS2024_95c6ae3f}
Maya Varma, Jean-Benoit Delbrouck, Zhihong Chen, Akshay Chaudhari, and Curtis Langlotz. 2024.
\newblock \href {https://proceedings.neurips.cc/paper_files/paper/2024/file/95c6ae3f3393786203a4b6dcb9df1036-Paper-Conference.pdf} {Ravl: Discovering and mitigating spurious correlations in fine-tuned vision-language models}.
\newblock In \emph{Advances in Neural Information Processing Systems}, volume~37, pages 82235--82264. Curran Associates, Inc.

\bibitem[{Wang et~al.(2024{\natexlab{a}})Wang, Jiang, Liu, Ma, Zhang, Pan, Liu, Gu, Xia, Li, Zhang, Wu, Liu, Zhong, Ge, Zhang, Qiang, Hu, Jiang, Zhang, Zhang, Shen, Liu, and Zhang}]{wang2024comprehensivereviewmultimodallarge}
Jiaqi Wang, Hanqi Jiang, Yiheng Liu, Chong Ma, Xu~Zhang, Yi~Pan, Mengyuan Liu, Peiran Gu, Sichen Xia, Wenjun Li, Yutong Zhang, Zihao Wu, Zhengliang Liu, Tianyang Zhong, Bao Ge, Tuo Zhang, Ning Qiang, Xintao Hu, Xi~Jiang, and 5 others. 2024{\natexlab{a}}.
\newblock \href {https://arxiv.org/abs/2408.01319} {A comprehensive review of multimodal large language models: Performance and challenges across different tasks}.
\newblock \emph{Preprint}, arXiv:2408.01319.

\bibitem[{Wang et~al.(2024{\natexlab{b}})Wang, Bai, Tan, Wang, Fan, Bai, Chen, Liu, Wang, Ge et~al.}]{wang2024qwen2}
Peng Wang, Shuai Bai, Sinan Tan, Shijie Wang, Zhihao Fan, Jinze Bai, Keqin Chen, Xuejing Liu, Jialin Wang, Wenbin Ge, and 1 others. 2024{\natexlab{b}}.
\newblock Qwen2-vl: Enhancing vision-language model's perception of the world at any resolution.
\newblock \emph{arXiv preprint arXiv:2409.12191}.

\bibitem[{Wang et~al.(2024{\natexlab{c}})Wang, Duan, and Cai}]{wang-etal-2024-multimodal}
Shuqi Wang, Xufeng Duan, and Zhenguang Cai. 2024{\natexlab{c}}.
\newblock \href {https://doi.org/10.18653/v1/2024.conll-1.32} {A multimodal large language model {\textquotedblleft}foresees{\textquotedblright} objects based on verb information but not gender}.
\newblock In \emph{Proceedings of the 28th Conference on Computational Natural Language Learning}, pages 435--441, Miami, FL, USA. Association for Computational Linguistics.

\bibitem[{Wang et~al.(2021)Wang, Wang, Shi, Li, Zhu, Liu, and Zhang}]{wang-etal-2021-counterfactual-adversarial}
Wei Wang, Boxin Wang, Ning Shi, Jinfeng Li, Bingyu Zhu, Xiangyu Liu, and Rong Zhang. 2021.
\newblock \href {https://doi.org/10.18653/v1/2021.findings-emnlp.413} {Counterfactual adversarial learning with representation interpolation}.
\newblock In \emph{Findings of the Association for Computational Linguistics: EMNLP 2021}, pages 4809--4820, Punta Cana, Dominican Republic. Association for Computational Linguistics.

\bibitem[{Weng et~al.(2024)Weng, Gao, Andrews, and Zhao}]{weng-etal-2024-images}
Zhaotian Weng, Zijun Gao, Jerone Andrews, and Jieyu Zhao. 2024.
\newblock \href {https://doi.org/10.18653/v1/2024.emnlp-main.878} {Images speak louder than words: Understanding and mitigating bias in vision-language model from a causal mediation perspective}.
\newblock In \emph{Proceedings of the 2024 Conference on Empirical Methods in Natural Language Processing}, pages 15669--15680, Miami, Florida, USA. Association for Computational Linguistics.

\bibitem[{Wolf et~al.(2020)Wolf, Debut, Sanh, Chaumond, Delangue, Moi, Cistac, Rault, Louf, Funtowicz, Davison, Shleifer, von Platen, Ma, Jernite, Plu, Xu, Le~Scao, Gugger, Drame, Lhoest, and Rush}]{wolf-etal-2020-transformers}
Thomas Wolf, Lysandre Debut, Victor Sanh, Julien Chaumond, Clement Delangue, Anthony Moi, Pierric Cistac, Tim Rault, Remi Louf, Morgan Funtowicz, Joe Davison, Sam Shleifer, Patrick von Platen, Clara Ma, Yacine Jernite, Julien Plu, Canwen Xu, Teven Le~Scao, Sylvain Gugger, and 3 others. 2020.
\newblock \href {https://doi.org/10.18653/v1/2020.emnlp-demos.6} {Transformers: State-of-the-art natural language processing}.
\newblock In \emph{Proceedings of the 2020 Conference on Empirical Methods in Natural Language Processing: System Demonstrations}, pages 38--45, Online. Association for Computational Linguistics.

\bibitem[{Wu et~al.(2024)Wu, Kuang, Zhu, Wang, Zheng, Han, Li, Chen, Wu, and Zhang}]{wu2024causalitylargelanguagemodels}
Anpeng Wu, Kun Kuang, Minqin Zhu, Yingrong Wang, Yujia Zheng, Kairong Han, Baohong Li, Guangyi Chen, Fei Wu, and Kun Zhang. 2024.
\newblock \href {https://arxiv.org/abs/2410.15319} {Causality for large language models}.
\newblock \emph{Preprint}, arXiv:2410.15319.

\bibitem[{Yang et~al.(2024{\natexlab{a}})Yang, Li, Xiao, Liu, Yang, Chen, Wang, Zhai, Li, and Zhang}]{10.1007/978-3-031-73636-0_27}
Dingkang Yang, Mingcheng Li, Dongling Xiao, Yang Liu, Kun Yang, Zhaoyu Chen, Yuzheng Wang, Peng Zhai, Ke~Li, and Lihua Zhang. 2024{\natexlab{a}}.
\newblock \href {https://doi.org/10.1007/978-3-031-73636-0_27} {Towards multimodal sentiment analysis debiasing via bias purification}.
\newblock In \emph{Computer Vision – ECCV 2024: 18th European Conference, Milan, Italy, September 29–October 4, 2024, Proceedings, Part LVIII}, page 464–481, Berlin, Heidelberg. Springer-Verlag.

\bibitem[{Yang et~al.(2024{\natexlab{b}})Yang, Yang, Li, Wang, Wang, and Zhang}]{Yang_2024_CVPR}
Dingkang Yang, Kun Yang, Mingcheng Li, Shunli Wang, Shuaibing Wang, and Lihua Zhang. 2024{\natexlab{b}}.
\newblock Robust emotion recognition in context debiasing.
\newblock In \emph{Proceedings of the IEEE/CVF Conference on Computer Vision and Pattern Recognition (CVPR)}, pages 12447--12457.

\bibitem[{Yang et~al.(2021{\natexlab{a}})Yang, Feng, Wang, and Zhang}]{9246699}
Xiaocui Yang, Shi Feng, Daling Wang, and Yifei Zhang. 2021{\natexlab{a}}.
\newblock \href {https://doi.org/10.1109/TMM.2020.3035277} {Image-text multimodal emotion classification via multi-view attentional network}.
\newblock \emph{IEEE Transactions on Multimedia}, 23:4014--4026.

\bibitem[{Yang et~al.(2021{\natexlab{b}})Yang, Feng, Zhang, and Wang}]{yang-etal-2021-multimodal}
Xiaocui Yang, Shi Feng, Yifei Zhang, and Daling Wang. 2021{\natexlab{b}}.
\newblock \href {https://doi.org/10.18653/v1/2021.acl-long.28} {Multimodal sentiment detection based on multi-channel graph neural networks}.
\newblock In \emph{Proceedings of the 59th Annual Meeting of the Association for Computational Linguistics and the 11th International Joint Conference on Natural Language Processing (Volume 1: Long Papers)}, pages 328--339, Online. Association for Computational Linguistics.

\bibitem[{Ye et~al.(2024)Ye, Zheng, Ma, Cao, Lai, Rehg, and Zhang}]{ye2024mmspubenchbetterunderstandingspurious}
Wenqian Ye, Guangtao Zheng, Yunsheng Ma, Xu~Cao, Bolin Lai, James~M. Rehg, and Aidong Zhang. 2024.
\newblock \href {https://arxiv.org/abs/2406.17126} {Mm-spubench: Towards better understanding of spurious biases in multimodal llms}.
\newblock \emph{Preprint}, arXiv:2406.17126.

\bibitem[{Yu et~al.(2024{\natexlab{a}})Yu, Yu, and Wang}]{yu2024api}
Runpeng Yu, Weihao Yu, and Xinchao Wang. 2024{\natexlab{a}}.
\newblock Api: Attention prompting on image for large vision-language models.

\bibitem[{Yu et~al.(2025)Yu, Zhang, Fu, Wu, Lu, Wang, Lu, Shen, Zhang, Song, Yan, Xu, Wen, Zhang, Huang, Wang, and Tan}]{yu2025aligningmultimodalllmhuman}
Tao Yu, Yi-Fan Zhang, Chaoyou Fu, Junkang Wu, Jinda Lu, Kun Wang, Xingyu Lu, Yunhang Shen, Guibin Zhang, Dingjie Song, Yibo Yan, Tianlong Xu, Qingsong Wen, Zhang Zhang, Yan Huang, Liang Wang, and Tieniu Tan. 2025.
\newblock \href {https://arxiv.org/abs/2503.14504} {Aligning multimodal llm with human preference: A survey}.
\newblock \emph{Preprint}, arXiv:2503.14504.

\bibitem[{Yu et~al.(2024{\natexlab{b}})Yu, Yao, Zhang, He, Han, Cui, Hu, Liu, Zheng, Sun, and Chua}]{yu2024rlhfv}
Tianyu Yu, Yuan Yao, Haoye Zhang, Taiwen He, Yifeng Han, Ganqu Cui, Jinyi Hu, Zhiyuan Liu, Hai-Tao Zheng, Maosong Sun, and Tat-Seng Chua. 2024{\natexlab{b}}.
\newblock Rlhf-v: Towards trustworthy mllms via behavior alignment from fine-grained correctional human feedback.
\newblock In \emph{Proceedings of the IEEE/CVF Conference on Computer Vision and Pattern Recognition (CVPR)}, pages 13807--13816.

\bibitem[{Zhang et~al.(2024{\natexlab{a}})Zhang, Ma, Guo, Li, Xu, Tang, and Hong}]{pmlr-v235-zhang24as}
Jie Zhang, Xiaosong Ma, Song Guo, Peng Li, Wenchao Xu, Xueyang Tang, and Zicong Hong. 2024{\natexlab{a}}.
\newblock \href {https://proceedings.mlr.press/v235/zhang24as.html} {Amend to alignment: Decoupled prompt tuning for mitigating spurious correlation in vision-language models}.
\newblock In \emph{Proceedings of the 41st International Conference on Machine Learning}, volume 235 of \emph{Proceedings of Machine Learning Research}, pages 59505--59519. PMLR.

\bibitem[{Zhang et~al.(2024{\natexlab{b}})Zhang, Yu, Wen, Wang, Zhang, Wang, Jin, and Tan}]{zhang2024debiasingmultimodallargelanguage}
Yi-Fan Zhang, Weichen Yu, Qingsong Wen, Xue Wang, Zhang Zhang, Liang Wang, Rong Jin, and Tieniu Tan. 2024{\natexlab{b}}.
\newblock \href {https://arxiv.org/abs/2403.05262} {Debiasing multimodal large language models}.
\newblock \emph{Preprint}, arXiv:2403.05262.

\bibitem[{Zhao et~al.(2024)Zhao, Si, Chen, Zhang, Sun, Zhang, and Chang}]{zhao2024lookingtextreducinglanguage}
Haozhe Zhao, Shuzheng Si, Liang Chen, Yichi Zhang, Maosong Sun, Mingjia Zhang, and Baobao Chang. 2024.
\newblock \href {https://arxiv.org/abs/2411.14279} {Looking beyond text: Reducing language bias in large vision-language models via multimodal dual-attention and soft-image guidance}.
\newblock \emph{Preprint}, arXiv:2411.14279.

\bibitem[{Zheng et~al.(2024{\natexlab{a}})Zheng, Wang, Zhou, and Zhang}]{zheng2024snse}
Guangmin Zheng, Jin Wang, Xiaobing Zhou, and Xuejie Zhang. 2024{\natexlab{a}}.
\newblock \href {https://arxiv.org/abs/2405.09848} {Enhancing semantics in multimodal chain of thought via soft negative sampling}.
\newblock \emph{Preprint}, arXiv:2405.09848.

\bibitem[{Zheng et~al.(2024{\natexlab{b}})Zheng, Zhang, Zhang, Ye, and Luo}]{zheng-etal-2024-llamafactory}
Yaowei Zheng, Richong Zhang, Junhao Zhang, Yanhan Ye, and Zheyan Luo. 2024{\natexlab{b}}.
\newblock \href {https://doi.org/10.18653/v1/2024.acl-demos.38} {{L}lama{F}actory: Unified efficient fine-tuning of 100+ language models}.
\newblock In \emph{Proceedings of the 62nd Annual Meeting of the Association for Computational Linguistics (Volume 3: System Demonstrations)}, pages 400--410, Bangkok, Thailand. Association for Computational Linguistics.

\bibitem[{Zhu et~al.(2024{\natexlab{a}})Zhu, Zhao, Ge, and Zhang}]{zhu2024vspa}
Ke~Zhu, Liang Zhao, Zheng Ge, and Xiangyu Zhang. 2024{\natexlab{a}}.
\newblock \href {https://doi.org/10.1145/3664647.3680993} {Self-supervised visual preference alignment}.
\newblock In \emph{Proceedings of the 32nd ACM International Conference on Multimedia}, MM '24, page 291–300, New York, NY, USA. Association for Computing Machinery.

\bibitem[{Zhu et~al.(2024{\natexlab{b}})Zhu, Zhuang, Zhang, Xu, Hu, Wu, and Zheng}]{ijcai2024p739}
Zhihong Zhu, Xianwei Zhuang, Yunyan Zhang, Derong Xu, Guimin Hu, Xian Wu, and Yefeng Zheng. 2024{\natexlab{b}}.
\newblock \href {https://doi.org/10.24963/ijcai.2024/739} {Tfcd: Towards multi-modal sarcasm detection via training-free counterfactual debiasing}.
\newblock In \emph{Proceedings of the Thirty-Third International Joint Conference on Artificial Intelligence, {IJCAI-24}}, pages 6687--6695. International Joint Conferences on Artificial Intelligence Organization.
\newblock Main Track.

\end{thebibliography}


\appendix
\label{sec:appendix}
\section{Causal Inference Theory and Counterfactual Causal Inference}
\begin{figure}[h]
    \centering
    \includegraphics[width=0.9\linewidth]{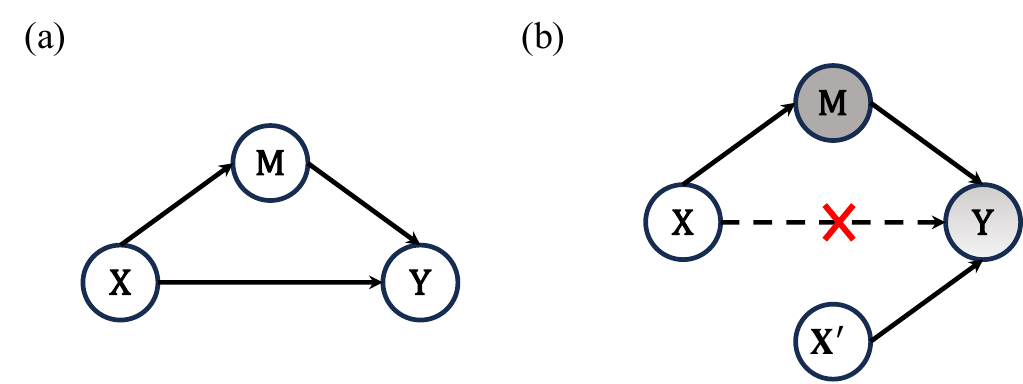}
    \caption{(a) An example of a causal graph, where $X$ transmits information to the outcome $Y$ via the mediator variable $M$. (b) The process of counterfactual intervention, observing the influence of $X$ on $Y$ via the mediator variable $M$ by interfering the link between $X$ and $Y$.}
    \label{fig:causal-graph}
\end{figure}
This section first introduces the theory of causal inference, which serves as the cornerstone for multimodal causal mediation analysis in Sec.~\ref{sec:multimodal-causal-graph}.

Causal graphs are highly generalized analytical tools used to reveal causal dependencies between variables. The prediction process of a model can be defined as a Directed Acyclic Graph (DAG) $G={V,E}$, where the set of nodes $V$ represents a series of intermediate factors involved in the prediction process, and the set of edges $E$ denotes the causal effects between them. Figure~\ref{fig:causal-graph} presents an example of a causal graph with three variables. In this graph, uppercase letters denote random variables, and lowercase letters denote their actual observed values. The causal relationship from a reference variable $X$ to an outcome variable $Y$ comprises two paths:
\begin{itemize}
\item $X\to Y$: The direct effect path, representing the direct effect of $X$ on $Y$.
\item $X\to M \to Y$: The indirect effect path, representing the effect of $X$ on $Y$ mediated by the intermediate variable $M$.
\end{itemize}

Counterfactual inference involves applying counterfactual treatment conditions to the model and observing the outcome of the original input under these new conditions, thereby deducing the influence of the counterfactual factors. This process can be formally expressed. The outcome for the original input is:
\begin{equation}
Y_{x,m}=Y(X=x,M=m),
\end{equation}
where $m=M_x=M(X=x)$. By applying a counterfactual treatment to $x$ to yield $x^*$, we obtain the counterfactual outcome:
\begin{equation}
Y_{x^*,M_{x^*}}=Y(X=x^*,M=M(X=x^*)),
\end{equation}
where $M_{x^*}$ represents the value of the mediator $M$ when $X$ is changed to $x^*$. Furthermore, a scenario can be constructed where X is counterfactually set to $x^*$, but the mediator $M$ retains the value it would have taken under the original input $x$. The outcome in this case is $Y_{x^*,M_x}$.

Causal effects quantify the difference between the outcome after intervening on a reference variable and the outcome in its natural (pre-intervention) state. According to causal theory, when the variable $X$ is changed from $x$ to $x^*$, the Total Effect (TE) on the model's outcome is defined as:
\begin{equation}
TE=\text{DIFF}(Y_{x,M_x}, Y_{x^*,M_{x^*}}),
\end{equation}
where DIFF represents a function that measures the difference between the predicted outcomes before and after the intervention. The Total Effect (TE) can be decomposed into the Natural Direct Effect (NDE) and the Total Indirect Effect (TIE). The NDE refers to the impact of a change in the causal variable $X$ (from $x$ to $x^*$) on the outcome $Y$ when the mediator variable $M$ is held at the level it would naturally assume if $X$ were counterfactually set: 
\begin{equation}
NDE=\text{DIFF}(Y_{x,M_{x^*}}, Y_{x^*,M_{x^*}}).
\end{equation}
This aims to isolate the impact of the reference variable on the outcome variable, holding the mediator constant at $M_{x^*}$.

The Total Indirect Effect (TIE) represents the effect of the reference variable $X$ indirectly influencing the outcome variable $Y$ through the mediator $M$. It is calculated by subtracting NDE from TE:
\begin{equation}
TIE=TE-NDE=\text{DIFF}(Y_{x,M_x}, Y_{x,M_{x^*}}).
\end{equation}
Thus, depending on the nature of the mediator $M$, NDE or TIE can be leveraged for unbiased estimation regarding the outcome variable Y:
\begin{itemize}
\item If the mediator $M$ is identified as a variable that potentially introduces bias and is not conducive to the final prediction. In this case, we hold $M$ at the value $M_{x^*}$, and observe the unbiased effect of $X$ on $Y$. This process is achieved through NDE.
\item If the mediator $M$ represents the pathway or features of interest, and only the effect mediated through $M$ is desired for an accurate prediction. Here, TIE can be used to calculate the indirect effect of $X$ on $Y$ via $M$, yielding an unbiased estimate of this mediated effect.
\end{itemize}

In this paper, we define M as the bias variable (representing spurious textual and visual context), and therefore, we employ NDE for unbiased estimation.

\begin{algorithm*}[!ht]
\caption{Counterfactual Image Generation via Attention Masks}
\label{alg:cf-image}
\begin{algorithmic}[1]
\Require
MLLM $model$, image $I$, prompt $P$, number of model layers $L$, image dimensions $H, W$, image patch grid dimensions $(patch_h, patch_w)$, enhancement factor $\alpha$
\Ensure counterfactual image $I_{cf}$
\State $Y_{out}\gets$ $model$.generate($I$, $P$) \Comment{Generate the neutral image analysis}
\State attns$\gets$ $model$.extract\_attentions($I$, $P$, $Y_{out}$) \Comment{Fetch attentions from $Y_{out}$ to $I$}
\State focused\_attns$\gets$ attns[L-4:L-1, :, :] \Comment{Select certain layers}
\State $\phi \gets$ \text{MeanPool}(focused\_attns) \Comment{Keep only final dimension only}
\State $\text{mask}_\text{2d}\gets$ Reshape($\phi$, ($patch_h$, $patch_w$)) \Comment{Reshape to 2 dimensional}
\State $\text{mask}_\text{norm}\gets$ Normalize($\text{mask}_\text{2d}$) \Comment{Normalization}
\State $\text{mask}_\text{enhanced}\gets\alpha\cdot\text{mask}_\text{norm}$ \Comment{Enhancement}
\State $\text{mask}_\text{smooth}\gets$ Conv2D($\text{mask}_\text{enhanced}$, kernel) \Comment{Smmothing}
\State $\text{mask}_\text{resized}\gets$ Interpolate($\text{mask}_\text{smooth}$, size=($H,W$)) 
\State $I_{cf}\gets$ AlphaBlend($I$, $\text{mask}_\text{resized}$) \Comment{Blending with original image}
\State\Return $I_{cf}$
\end{algorithmic}
\end{algorithm*}
\section{Counterfactual Content Construction}
\label{app:counterfactual_content_construction}
Guided by the multimodal causal mediation framework, we explicitly construct counterfactual samples that isolate textual and visual spurious contexts. Existing counterfactual generation approaches typically rely on rule-based heuristics, manipulation of latent representations or pre-existing annotations in datasets, and thus lack the granularity and flexibility for our detailed causal framework. In contrast, we propose leveraging large MLLMs to automate counterfactual input generation at the raw input level.

\subsection{Counterfactual Text Input Construction}
For textual inputs, we aim to clearly distinguish between key semantic content ($T_{semantic}$) and spurious context which is potentially bias-inducing ($T_{spurious}$). The counterfactual textual input is generated through the following procedure:

\paragraph{Identifying Semantic Content via Prompting.}
We prompt a large language model to extract primary semantic content relevant to the prediction task. We avoid directly identifying spurious context since irrelevant segments often lack stable semantic indicators, causing imprecise model responses. Instead, extracting key semantic segments, which inherently possess clear and consistent semantic features, provides a more precise foundation, allowing us to reliably obtain spurious context by subtracting the identified semantic content from the original text.

\paragraph{Generating Context-only Counterfactuals.}
Once critical semantic components are identified, we replace these segments with a neutral placeholder token (\texttt{[MASK]}), ensuring grammatical fluency while preserving only the bias-prone content:
\begin{equation}
T_{spurious} = T \backslash T_{semantic}
\end{equation}

\subsection{Counterfactual Image Input Construction}
Analogous to the textual scenario, we need to isolate visual content relevant to the prediction task ($I_{semantic}$) from spurious visual contexts ($I_{spurious}$). Due to the complexity of accurately distinguishing these regions, we propose an automated approach based on attention mechanisms from MLLM, enabling fine-grained visual counterfactual generation.

\paragraph{Extracting Visual Attention via Neutral Prompts.}
Analogous to our textual approach, we prompt the vision-language model with neutral descriptions rather than task-specific instructions. Unlike prior methods, which often rely on coarse or manual visual annotations, our method proactively extracts fine-grained visual content critical to the prediction task. Motivated by \citet{yu2024api}, task-specific prompts would introduce semantic biases related to internal task priors, potentially distorting the saliency distribution. In contrast, neutral prompting allows the model to identify image regions purely based on intrinsic visual importance.

Formally, given textual tokens $X_{text}$ and image patches $X_{image}$, the model generates descriptive tokens $Y_{out}$. Attention scores from $Y_{out}$ to $X_{image}$ are extracted from deeper transformer layers (e.g., layers 29–31 of the 32-layer Qwen2-VL-7B) to capture richer, more fine-grained multimodal interactions~\citep{wang-etal-2024-multimodal}. The attention score for the image patch at spatial position $(i, j)$ is computed as:
\begin{equation}
\phi_{i,j}=\sum_{h,m} A_{m,t}^{h}, \quad t=j+P\cdot(i-1).
\end{equation}
Here, $A_{m,t}^{h}$ denotes attention weights from the $m$-th token of the generated sequence to the $t$-th image patch across head $h$, and $P$ denotes the number of patches per row in the patch sequence.

\paragraph{Mask Enhancement for Counterfactual Image Generation}
To generate visually coherent and semantically meaningful counterfactual images from the raw attention masks, we apply several post-processing steps. First, the attention scores are normalized into a standardized range [0, 1] to stabilize mask values and facilitate subsequent manipulation. Second, we enhance the contrast of these normalized masks by scaling, thus emphasizing visually salient regions and clearly distinguishing critical visual elements from background noise. Third, a spatial smoothing operation (convolutional filtering) is performed to prevent abrupt transitions between masked and unmasked regions, maintaining visual continuity. Finally, we interpolate the smoothed masks back to the original image dimensions and blend them seamlessly with the original images via alpha blending. This procedure ensures that the resulting counterfactual images preserve contextual coherence while effectively highlighting the identified visual regions. Alg.~\ref{alg:cf-image} summarizes this detailed workflow precisely.

\subsection{Quality Control via Router}
While the proposed automated approach efficiently constructs fine-grained counterfactual inputs, the inherent complexity and variability of multimodal content might occasionally lead to suboptimal or noisy counterfactual samples. To mitigate potential quality issues arising from these automatically generated inputs, we incorporate a router mechanism that dynamically assesses their suitability for subsequent causal analysis. Specifically, the router determines whether each generated counterfactual input should be utilized or discarded based on its semantic reliability and consistency. Further details regarding the design and operational logic of the router are elaborated in later sections.

\section{Detailed Experimental Setup}
\label{app:experimental_details}
\begin{table*}[!ht]
\small
\centering
\begin{tabular}{@{}lccccc@{}}
\toprule
 & \multicolumn{2}{c}{\textbf{MMSD2.0}} & \multicolumn{3}{c}{\textbf{MVSA-Multi}} \\ \cmidrule(l){2-6} 
 & \#\textbf{Sarcasm} & \#\textbf{Non-sar} & \#\textbf{Positive} & \#\textbf{Neutral} & \#\textbf{Negative} \\ \midrule
\textbf{Train} & 8642 & 11174 & 8954 & 3461 & 1023 \\
\textbf{Valid} & 959 & 1451 & 1186 & 483 & 124 \\
\textbf{Test} & 959 & 1450 & 1177 & 463 & 151 \\ \bottomrule
\end{tabular}
\caption{Statistics of dataset MMSD2.0 and MVSA-Multi}
\label{tab:data-stats}
\end{table*}

\subsection{Datasets and Evaluation Metrics}
\label{app:datasets_metrics}
Our experiments were conducted on two distinct multimodal datasets to address sarcasm detection and sentiment analysis. Statistical details for both datasets are presented in Tab.~\ref{tab:data-stats}. For multimodal sarcasm detection, we utilized the MMSD2.0 dataset~\citep{qin-etal-2023-mmsd2}. We directly adopted the official dataset partitions for training, validation, and testing as provided by the original authors. For multimodal sentiment analysis, we employed the MVSA-Multi subset of the MVSA dataset~\citep{MVSA}. MVSA is a widely recognized benchmark in this domain, constructed from Twitter posts. The MVSA dataset consists of two subsets: MVSA-Single (MVSA-S), which contains 5,129 samples, each annotated by a single annotator, and MVSA-Multi (MVSA-M), which includes 19,600 samples, with each sample receiving three independent annotations. We selected MVSA-Multi for our experiments due to its superior annotation reliability and lower noise levels, thereby supporting more robust experimental findings. For the MVSA-Multi dataset, we randomly partitioned the data into training, validation, and testing sets, adhering to a 7.5:1:1 ratio.

To comprehensively assess model efficacy, we used specific evaluation metrics for each task. For sarcasm detection (MMSD2.0), where "Non-sar" instances outnumber "Sarcasm," metrics like Precision, Recall, and F1-score were used alongside Accuracy to ensure the model's ability to identify the minority sarcastic class wasn't obscured. Similarly, for the more significantly imbalanced MVSA-Multi sentiment dataset (with "Negative" as a clear minority), Accuracy was supplemented with Macro-F1 and Weighted-F1 to provide a fairer assessment of performance across all classes

\subsection{Implementation Details}
Our experiments utilized two powerful large multimodal model series, Qwen2-VL~\citep{wang2024qwen2} and InternVL2.5~\citep{chen2025expandingperformanceboundariesopensource}, specifically the Qwen2-VL-7B and InternVL2.5-4B versions, trainable on a single NVIDIA A100 80G GPU. We employed LoRA~\citep{hu2022lora} for fine-tuning, using the LLaMA-Factory framework~\citep{zheng-etal-2024-llamafactory} for Qwen2-VL-7B (with data in dialogue format) and official scripts for InternVL2.5-4B. Key LoRA parameters included a rank of 16, 10 epochs, a learning rate of 4e-5, weight decay of 0.01, a batch size of 8, and 2 gradient accumulation steps, accelerated with DeepSpeed Stage 1~\citep{rajbhandari2020zeromemoryoptimizationstraining}. Checkpoints were saved every 500 steps, with the best model selected based on minimum validation loss. Each expert training requires 1 day around.

We choose CLIP~\citep{pmlr-v139-radford21a} to serve as the backbone for our router model. This Router was trained for 15 epochs using the Transformers library~\citep{wolf-etal-2020-transformers}, with the best checkpoint chosen based on the F-0.5 score on the validation set to prioritize precision in identifying samples needing correction. Optimal hyperparameters for the inference-time debiasing process were identified using Bayesian optimization~\citep{NIPS2012_05311655} via the skopt package, employing a Gaussian process model with a maximum of 50 function evaluations. To ensure the robustness and reliability of our experimental results, we conducted each proposed method five times independently and reported the average performance across these trials.

\subsection{Debiasing Category Distribution}
For completeness, we summarize the training-set distribution of debiasing categories (InternVL2.5). The majority of samples are labeled \emph{None}, yet a non-trivial portion requires single- or dual-modality debiasing, motivating adaptive expert routing.

\begin{table*}[h]
\centering
\small
\begin{tabular}{lcccc}
\toprule
\textbf{Dataset / Category} & \textbf{None} & \textbf{Image} & \textbf{Text} & \textbf{Both} \\
\midrule
MMSD2.0 (n=19{,}816)   & 86.8\% & 4.2\% & 7.0\% & 2.0\% \\
MVSA-Multi (n=13{,}438) & 93.2\% & 2.5\% & 2.1\% & 2.2\% \\
\bottomrule
\end{tabular}
\caption{Summary of debiasing-category prevalence (percentage of training set).}
\label{tab:debias-dist-appendix}
\end{table*}

\begin{figure*}[!ht]
\centering
  \includegraphics[width=0.45\linewidth]{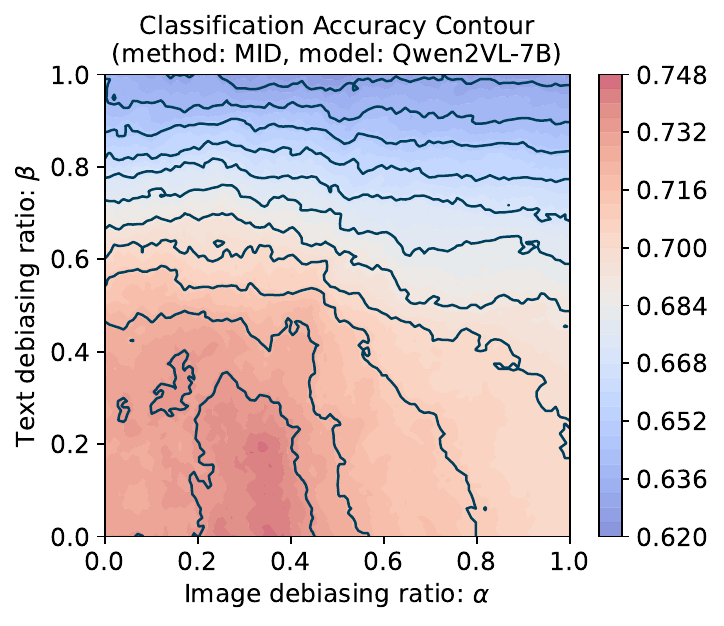} 
  \includegraphics[width=0.45\linewidth]{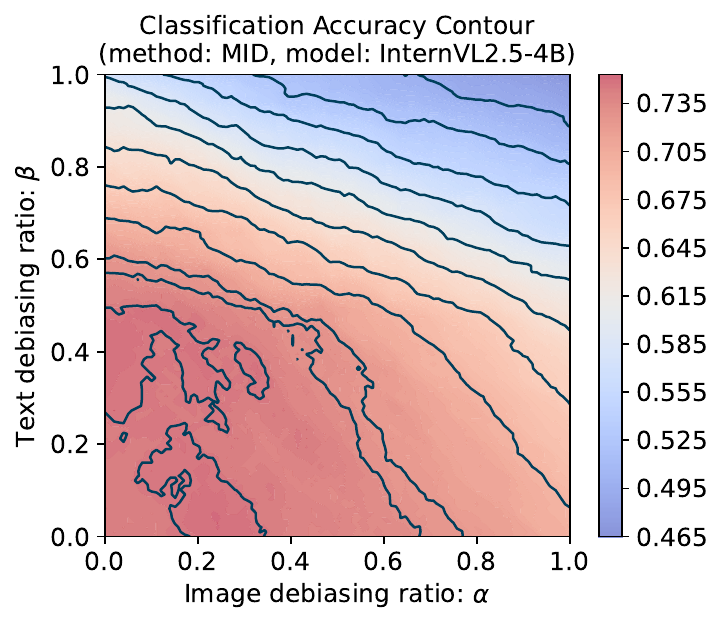}
  \caption{Contour plots illustrating the influence of image debiasing ratio ($\alpha$) and text debiasing ratio ($\beta$) on classification accuracy when applying MID to Qwen2-VL-7B and InternVL2.5-4B models (vanilla model without training).}
  \label{fig:hyperparam-contour}
\end{figure*}
\subsection{Comparing Methods}
\label{appendix:comparing_methods}
In order to comprehensively evaluate the performance of our proposed methods, we considered several settings and selected representative methods for comparison: 

(1) Methods specific for downstream tasks. For multimodal sarcasm detection, we choose \textbf{HFM}~\citep{cai-etal-2019-multi}, which introduced a hierarchical fusion approach for multi-modal sarcasm detection, distinctively treating image attributes as a third modality alongside text and image features; \textbf{Attn-BERT}~\citep{pan-etal-2020-modeling}, which applied BERT-based architectures with a significant attention mechanism to model user expressions in multi-modal content; \textbf{CMGCN}~\citep{liang-etal-2022-multi}, constructing of an instance-specific cross-modal graph to explicitly map relationships between image objects and textual words. It then employed a graph convolutional network to learn and identify incongruity within these structures; \textbf{HKE}~\citep{liu-etal-2022-towards-multi-modal} focused on advancing representation learning for heterogeneous knowledge graphs, embed diverse types of entities and relations, thus capturing complex semantics; \textbf{Multi-view CLIP}~\citep{qin-etal-2023-mmsd2} utilized CLIP by processing information from multiple views (text, image, and their interaction) to capture multi-grained cues.

For multimodal sentiment analysis, we choose \textbf{MVAN}~\citep{9246699}, which introduced a multi-view attention mechanism that captures image features from both object and scene perspectives, and employed a memory network that is continually updated to obtain deep semantic features; \textbf{MGNNS}~\citep{yang-etal-2021-multimodal}, constructed separate graphs for text and image modalities to capture the global co-occurrence characteristics of the dataset and utilized multi-channel graph neural networks to learn multimodal representations with a multi-head attention mechanism for in-depth fusion; \textbf{CLMLF}~\citep{li-etal-2022-clmlf} combined contrastive learning with a multi-layer fusion strategy to help the model learn common sentiment-related features across modalities; \textbf{MDSE}~\citep{10445820} focused on identifying sentiment expressions specific to individual modalities by leveraging semi-supervised variational autoencoders.

(2) Methods for multimodal causal debias.
\textbf{TFCD}~\citep{ijcai2024p739} targeted at biases in multi-modal sarcasm detection, specifically the model's over-reliance on frequently occurring non-sarcastic words and static co-occurrences between training data labels and modal features. The challenge of non-sarcastic word bias in the textual modality, as identified by TFCD, aligns with our the text branch in our proposed causal graph.
\textbf{MCIS}~\citep{10.1007/978-3-031-73636-0_27} designed for multi-modal sentiment analysis and explicitly differentiates between two common types of biases: utterance-level label bias and word-level context bias. The approach adopted by MCIS shares significant similarities with TFCD, with the primary distinction being the application task. 
\textbf{CF-MSA}~\citep{chen2024multimodalsentimentanalysisbased} addressed biases in both textual and visual modalities for multi-modal sentiment analysis. While CF-MSA shares our focus on addressing biases in both modalities, it is implemented based on BERT and operates at the feature representation level, and estimated unimodal bias by completely masking one modality.

\section{MRID: Multimodal Router-Guided Inference Debiasing}
\label{app:mrid}
To further analyze the effectiveness of the dynamic routing mechanism from MME-JD in an inference-only setting, we introduce Multimodal Router-Guided Inference Debiasing (MRID). This approach adapts the standard MID framework by incorporating the router to selectively apply modality-specific debiasing, rather than uniformly correcting for both modalities.

Specifically, following Sec.~\ref{sec:mid}, we obtain outputs under three scenarios: original prediction ($p_0$), text-spurious prediction $p_t$ and image-spurious prediction ($p_i$). The MME-JD router (as described in Section~\ref{sec:router}), which was trained to predict an optimal expert strategy is then employed. The router takes the set of inputs $(i, t, \hat{i}, \hat{t})$and outputs a strategy $c^*$. Based on the router's decision $c^*$, the final debiased prediction $\tilde{p}$ for MRID is computed conditionally:
\begin{align}
\tilde{p} \;=\;
\begin{cases}
p_0, &  c^*=0 \\
p_0 - \alpha_1 \cdot p_i, &  c^*=1 \\
p_0 - \beta_2 \cdot p_t, &  c^*=2 \\
p_0 - \alpha_3 \cdot p_i - \beta_3 \cdot p_t. &  c^*=3
\end{cases}
\label{eq:mrid}
\end{align}
The hyperparameters $\alpha_{c}, \beta_{c}$ are searched on validation set as Sec.~\ref{sec:mid}.

\section{Analysis on Hyperparam for MID}
To examine the usage of linear mode inference-time debaising, we analyzed the classification accuracy sensitivity to the coefficients $\alpha$ (text debias degree) and $\beta$ (image debias degree) in MID. The contour plots presented in Fig.~\ref{fig:hyperparam-contour}. For both models, the resulting accuracy is clearly dependent on the specific choices of $\alpha, \beta$. It is evident that optimal performance is typically achieved when both $\alpha, \beta$ are non-zero. For the Qwen2VL-7B, peak accuracy is observed around (0.4, 0.2), while the highest accuracy for InternVL2.5-4B is achieved in $\alpha \in [0.1, 0.4], \beta \in [0.3, 0.5]$. Setting these coefficients to extreme values (1.0) can lead to suboptimal performance, potentially due to over-correction which might suppress true signal along with bias. The slightly different optimal regions and peak accuracies between the two base models also suggest that these hyperparameters may benefit from model-specific tuning for best results.

\section{Case Study}
\label{app:case_study}
Below we present two samples selected from training set to illustrate the usage of multimodal debias.

\begin{figure*}[h!]
    \centering
    \begin{subfigure}{0.48\textwidth}
        \centering
        \includegraphics[width=\linewidth]{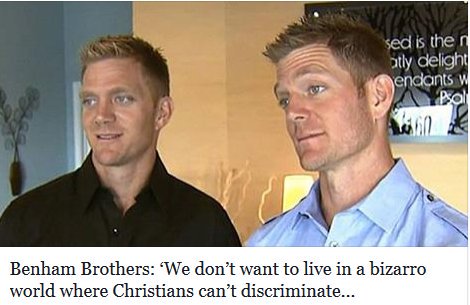} 
        \caption{Origin Text: nothing says equality like discrimination}
        \label{fig:case1_original}
    \end{subfigure}
    \hfill 
    \begin{subfigure}{0.48\textwidth}
        \centering
        \includegraphics[width=\linewidth]{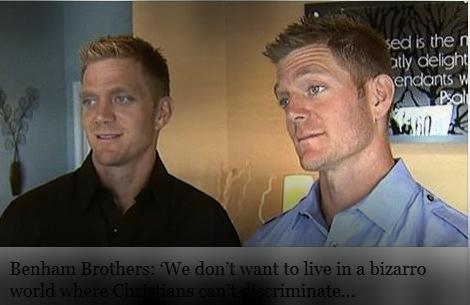}
        \caption{Masked Text: nothing says [MASK] like [MASK]}
        \label{fig:case1_masked}
    \end{subfigure}
    \begin{subfigure}{0.48\textwidth}
        \centering
        \includegraphics[width=\linewidth]{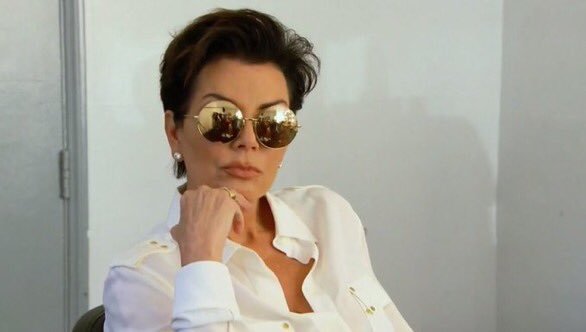}
        \caption{Origin Text: before u say jongin 's " extra " and " too much " , ask urself ... are ur favs even " enough " ?}
        \label{fig:case2_original}
    \end{subfigure}
    \hfill
    \begin{subfigure}{0.48\textwidth}
        \centering
        \includegraphics[width=\linewidth]{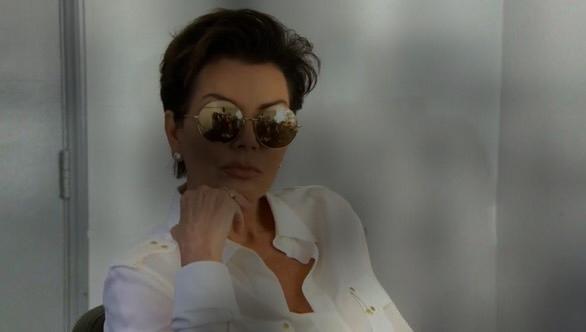}
        \caption{Masked Text: before u say jongin 's " [MASK] " and " [MASK] " , ask urself ... are ur favs even " [MASK] " ?}
        \label{fig:case2_masked}
    \end{subfigure}
    \caption{The original samples versus the masked samples (image and text inputs after masking major semantic content) with core semantic and contextual information removed.}
    \label{fig:case_study}
\end{figure*}

\begin{table*}[h]
\centering
\small
\begin{tabular}{lccc}
\toprule
\textbf{Case} & \textbf{Ground Truth} & \textbf{Before Debias} & \textbf{After Debias} \\
\midrule
Fig.~\ref{fig:case_study} a,b & sarcastic & [0.21, \textbf{0.79}] & [0.01, \textbf{0.99}] \\
Fig.~\ref{fig:case_study} c,d & non-sarcastic & [0.47, \textbf{0.53}] & [\textbf{0.72}, 0.28] \\
\bottomrule
\end{tabular}
\caption{Probability distributions over \{non-sarcastic, sarcastic\} before and after debiasing.}
\label{tab:case-probs}
\end{table*}

\paragraph{Enhancing judgement by removing multimodal bias}
In case 1 (Fig.~\ref{fig:case_study}(a–b)), the caption “nothing says equality like discrimination” expresses sarcasm through a polarity-incongruent pairing and a fixed “nothing-says-X-like-Y” template. On the text side we mask the trigger content words \emph{equality} and \emph{discrimination}; on the image side we occlude the headline strip and adjacent high-saliency area, leaving a neutral studio background. The model’s output changes from $[0.21,\,0.79]$ (non-sarcastic, sarcastic) to $[0.01,\,0.99]$. This shift indicates that a non-trivial portion of the initial non-sarcastic probability was supported by background cues rather than the ironic semantics, and that discounting background-only evidence yields a more calibrated prediction without altering the correct class.

\paragraph{Correcting judgement by mitigating textual bias}
In case 2 (Fig.~\ref{fig:case_study}(c–d)), the ground-truth label is non-sarcastic, yet the original prediction is borderline ($[0.47,\,0.53]$). Here the dominant source of spurious evidence lies in the text: quoted evaluatives such as “extra,” “too much,” and “enough” are frequent correlates of sarcasm in web corpora and thus behave as prior-driven indicators. We mask these tokens (and lightly occlude the highest-saliency facial region in the image), which removes much of the sarcasm prior while preserving the author’s non-ironic intent. The distribution moves to $[0.72,\,0.28]$, aligning with the label and illustrating that attenuating text-side priors is effective when surface lexical markers, rather than semantics, are responsible for the erroneous bias.

\section{Prompt Templates}
\begin{tcolorbox}[
    colback=lightgrassgreen, 
    colframe=black!60!white,   
    title=Image Analysis Prompt, 
    fonttitle=\bfseries,
    fontupper=\small
]
You are provided with an image from a tweet with the associated text: ``\%s''. Analyze the image and categorize its visual elements based on their semantic relevance:
\begin{itemize}
    \item \textbf{Main Content Elements:} Identify visual elements in the image that provide meaningful semantic clues, such as emotionally charged objects, facial expressions, or thematic components. These elements should align with or contradict the textual information.
    \item \textbf{Context Elements:} Identify generic, non-essential visual details (e.g., background patterns, irrelevant objects) that do not contribute significant semantic value to understanding the text-image relationship.
\end{itemize}
\textbf{Output Format:}
\begin{itemize}
    \item \textbf{Main Content Elements:} [List of visual elements]
    \item \textbf{Context Elements:} [List of visual elements]
\end{itemize}
\textbf{Analysis Process:} Provide a brief explanation of how the main content elements were identified and their connection (or contradiction) with the text. Highlight how the context elements were separated based on their lack of semantic importance.
\end{tcolorbox}
\begin{tcolorbox}[
    colback=lightblue, 
    colframe=black!60!white,   
    title=Text Analysis Prompt, 
    fonttitle=\bfseries,
    fontupper=\small
]
You are provided with an image from a tweet with the associated text: ``\%s''. Identify and categorize the words in the text based on their semantic relevance:
\begin{itemize}
    \item \textbf{Main Content Words:} Extract words or phrases that provide meaningful semantic clues, such as emotional, thematic, or descriptive elements.
    \item \textbf{Context Words:} Extract words or phrases that are generic, stylistic, or non-essential (e.g., stop words, filler adjectives) and do not contribute significant semantic value.
\end{itemize}
\textbf{Output Format:}
\begin{itemize}
    \item \textbf{Analysis Process:} Provide a brief explanation of how the main content words were identified and their relation with the image. Highlight how the context words were separated based on their lack of semantic importance.
    \item \textbf{Main Content Words:} [List of words/phrases]
    \item \textbf{Context Words:} [List of words/phrases]
\end{itemize}
\end{tcolorbox}

\end{document}